\documentclass[preprint,12pt,3p]{elsarticle}

\usepackage{longtable}
\usepackage{amssymb}
\usepackage{amsmath}
\usepackage{algorithm, eqparbox, array} 
\usepackage{algcompatible}
\usepackage{algpseudocode}
\usepackage{hyperref}
\usepackage{color}
\usepackage{rotating}

\algnewcommand{\LeftComment}[1]{\Statex \(\triangleright\) #1}

\hypersetup{
    colorlinks=true,
    linkcolor=blue,
    filecolor=blue,      
    urlcolor=blue,
    citecolor=blue,
}

\journal{Elsevier}

\begin{document}

\begin{frontmatter}

\title{Ensembles of Localised Models for Time Series Forecasting}

\cortext[cor1]{Corresponding author. Postal Address: Faculty of Information Technology, P.O. Box 63 Monash University, Victoria
3800, Australia. E-mail address: rakshitha.godahewa@monash.edu}

\author[FIT]{Rakshitha Godahewa \corref{cor1}}
\author[MEL]{Kasun Bandara}
\author[FIT]{Geoffrey I.\ Webb}
\author[FB]{Slawek Smyl}
\author[FIT]{Christoph Bergmeir}

\address[FIT]{Department of Data Science and Artificial Intelligence, Faculty of Information Technology, Monash University, Melbourne, Australia}
\address[MEL]{School of Computing and Information Systems, Melbourne Centre for Data Science, University of
Melbourne, Melbourne, Australia.}
\address[FB]{Uber Technologies Inc, San Francisco, California, United States}

\begin{abstract}%
With large quantities of data typically available nowadays, forecasting models that are trained across sets of time series, known as Global Forecasting Models (GFM), are regularly outperforming traditional univariate forecasting models that work on isolated series. As GFMs usually share the same set of parameters across all time series, they often have the problem of not being localised enough to a particular series, especially in situations where datasets are heterogeneous. We study how ensembling techniques can be used with generic GFMs and univariate models to solve this issue. Our work systematises and compares relevant current approaches, namely clustering series and training separate submodels per cluster, the so-called ensemble of specialists approach, and building heterogeneous ensembles of global and local models. We fill some gaps in the existing GFM localisation approaches, in particular by incorporating varied clustering techniques such as feature-based clustering, distance-based clustering and random clustering, and generalise them to use different underlying GFM model types. 
We then propose a new methodology of clustered ensembles where we train multiple GFMs on different clusters of series, obtained by changing the number of clusters and cluster seeds.
Using Feed-forward Neural Networks, Recurrent Neural Networks, and Pooled Regression models as the underlying GFMs, in our evaluation on eight publicly available datasets, the proposed models are able to achieve significantly higher accuracy than baseline GFM models and univariate forecasting methods.
\end{abstract}


\end{frontmatter}

\section{Introduction}
\label{sec:intro}

Nowadays, many businesses and industries, such as retail, energy, and tourism, routinely collect massive amounts of time series from similar sources that require accurate forecasts for better decision making and strategic planning. Recently, global forecasting models (GFM) that build a single model across many time series \citep{ref_105}, have shown their potential in providing accurate forecasts under these circumstances, e.g., by winning the prestigious M4 and M5 forecasting competitions~\citep{ref_30,makridakis_2020_m5}. Compared with traditional univariate forecasting methods, such as Exponential Smoothing~\citep[ETS,][]{ref_112} and Auto-Regressive Integrated Moving Average models~\citep[ARIMA,][]{ref_113} that build separate models to forecast each series, GFMs are built using all the available time series, and are capable of exploiting cross-series information and control overall model complexity better through shared parameters~\citep{pablo_2020_principles}, though their complexity is often required to be controlled based on heterogeneity and amount of the training data available~\citep{hewamalage_2020_simulation}. 

A GFM in its purest form, e.g., a pooled regression model~\citep[PR, ][]{ref_114,gelman2007data}, shares all parameters across all series, and can therewith be seen as an extreme opposite case of a traditional univariate forecasting model that has only per-series parameters. Such GFMs now consequently often have the problem that they are not localised enough for particular series, especially when the time series datasets are heterogeneous. To address this issue, researchers have developed special models with both global parameters shared across all series and parameters per each series \citep{ref_1, sen_2019_think}. 
However, another more generally applicable and generic approach to address this problem that can be used with any GFM is to build models across subsets of the overall time series database. This strategy allows both to control model complexity and to build more localised models, focused on certain groups of time series. The challenge then is how to define these subgroups of series and how to combine the outputs of the resulting submodels. 

Clustering is one approach that can be used, both as a way to make the overall model more complex in a controlled way and to address data heterogeneity. 
\citet{pablo_2020_principles} show that random clustering can improve forecasting accuracy, as it increases overall model complexity. 
To address data heterogeneity, \citet{ref_2} propose an approach that initially clusters the time series based on a similarity measure, and then trains separate GFMs on each identified cluster. 
Similarly, clusters can be defined using domain knowledge, for example by using spatial proximity in applications such as environmental prediction and weather prediction \citep{lerch2016similarity}. 
The winning method of the M4 forecasting competition \citep{ref_1} uses an ensembling approach known as \emph{ensemble of specialists}. This approach trains a group of GFMs that specialise by being more likely to train on series they already perform better on than the other models in the ensemble.
Yet another approach is to combine global and local models in a forecast combination approach, as in~\cite{godahewa_2020_weekly}.

The first main contribution of our study is to systematically compare the different generic approaches of localising GFMs proposed in the literature, namely clustering, ensemble of specialists, and forecast combination of local and global models. Therefore, we generalise the ensemble of specialists and clustering approaches from the literature, using them with different baseline GFM architectures.
Also, the current work in the literature only uses random and feature-based clustering approaches for subgrouping the time series. We fill this gap by additionally using distance-based clustering. 

In machine learning, ensembling techniques are used very successfully to reduce model variance and model bias, and to quantify the uncertainty by aggregating predictions over multiple models~\citep[see, e.g., ][]{schapire_1999_adaboost,breiman_2001_random}. But also in the field of forecasting, it is well established for decades that forecast combinations lead to improved accuracy~\citep{granger_1969_combination,ref_58}. Forecast combinations address the limitations following from the \textit{No Free Lunch Theorem}  by \citet{wolpert2002lunch} which states that there is no single best prediction algorithm that is suitable for any application. Forecast combinations containing diverse models \citep{brown2005managing} incorporate the strengths of different models and hence, they are capable of providing accurate forecasts for many applications compared with single forecasting models.
 
This motivates us to propose an ensembling approach to localise GFMs which is the second main contribution of this study. Our proposed ensemble models contain localised GFMs trained over subsets of collections of time series with different model complexities. In particular, we first subgroup the time series using two feature-based clustering approaches: K-means \citep{ref_90} and K-means++ \citep{ref_91}, and a distance-based clustering approach, K-medoids \citep{ref_95} with Dynamic Time Warping (DTW) distance in multiple iterations by either varying the number of clusters or the cluster seed. Then, multiple GFMs are trained per each identified subgroup, for multiple iterations, generating multiple forecasts for each time series. These forecasts are then averaged to obtain the final predictions. 
Here, the issue of information loss in traditional clustering-based GFM approaches, as series are assigned to either one cluster or another and only one model is built per cluster, is addressed by building multiple GFMs over different clusters. We also show that more sophisticated clustering is a good way to add model complexity in a directed and focused way, especially if the amount of data is limited. 
Furthermore, we quantify in our extensive experimental study how localisation and increasing model complexity affect the performance of GFMs.
As the primary forecasting modules of our framework, we use linear and non-linear GFM models, namely: PR models, Feed-Forward Neural Networks (FFNN), and Recurrent Neural Networks (RNN). All implementations of this study are publicly available at: \url{https://github.com/rakshitha123/Localised_Ensembles}.

The remainder of this paper is organized as follows: Section \ref{sec:related_work} reviews the related work. Section \ref{sec:methods} introduces the problem statement and the methods used for the experiments such as FFNNs, RNNs with Long Short-Term Memory (LSTM) cells, PR models, and ensemble and non-ensemble models. Section \ref{sec:framework} explains the experimental framework, including the datasets, hyperparameter tuning, error metrics and benchmarks. We present an analysis of the results in Section \ref{sec:results}. Section \ref{sec:conclusion} concludes the paper and discusses possible future research.

\section{Related Work}
\label{sec:related_work}

In the following, we discuss related work and the theoretical basis in the areas of global models, localisation of global models, ensembling, and time series clustering.

\subsection{Global Models in Forecasting}

GFMs \citep{ref_105} are a relatively recent trend in forecasting and have become popular after being the base of the winning solution of the M4 forecasting competition \citep{ref_1}. 
Usually, GFMs build a single model with a set of global parameters across many series. In contrast to local models, they are capable of learning the cross-series information during model training with a smaller amount of parameters.
\citet{pablo_2020_principles} show that the complexity of local models increases with the number of series as a separate model is fitted for each series whereas the complexity of global models remains the same regardless of the amount of series.
Those authors also prove that for a particular set of series, there always exists a global model that can produce the same forecasts as given by a collection of local models. Thus, in principle GFMs are applicable to any forecasting problem, not only in situations where series are related, as in earlier work typically postulated \cite{hewamalage_2020_simulation}. However, the proof does not provide a way to construct such global models, so that in practice, there are still complex interdependencies between the degree of relatedness of series, the amount of data available, and the complexity of the models \cite{hewamalage_2020_simulation}. \citet{hewamalage_2020_simulation} perform a simulation study to characterise these interdependencies, and define relatedness using a model-based definition, in the way that series are defined to be related if their data generating processes are similar.

In terms of particular GFMs proposed in the literature, in an early work, \citet{duncan_2001_forecasting} use a Bayesian pooling approach to extract information from similar time series to train GFMs.
A linear auto-regressive GFM trained across a set of time series is known as a PR model, as discussed by \citet{ref_114}.  
\citet{ref_23} use global RNNs for forecasting.
\citet{ref_99} propose a global auto-regressive RNN named DeepAR for probabilistic forecasting.
Since 2018, many more models have been proposed in this space, e.g., \citet{oreshkin_2019_nbeats} propose the forecasting framework N-BEATS which trains deep neural networks incorporating cross-series information, and~\citet{ref_6} present an overview of globally trained RNNs. The M5 forecasting competition~\citep{makridakis_2020_m5} was won by a globally trained gradient-boosted tree, and also all recent Kaggle competitions in forecasting were won by global models~\citep{BOJER2020}.

Though GFMs may be relatively new in their appreciation in forecasting, similar concepts have been used in other fields, most prominently pooled regression models \cite{gelman2007data}, e.g., in the social sciences, and sub-population models in medicine \cite{kent_2007_limitations}.
\citet{gelman2007data} argue that developing prediction models with pooling and without pooling data are extreme opposite cases where both approaches have their own issues. The no-pooling approach tends to overfit the data, especially for instances with a smaller amount of training data. On the other hand, pooling or a global approach may underfit the data and tends to provide averaged predictions for all instances and therewith it will ignore specifics of individual instances (see also \citet{kent_2007_limitations}). Thus, those authors argue that the middle ground between pooling and no-pooling approaches is often more desirable, which in their context is called \textit{multilevel analysis}. In line with this concept, in our work, we use a middle ground between global (pooling) and local (no-pooling) approaches where we train global models across sets of time series, ensuring that the models are adequately localised by using more appropriate time series groups to train them.

\subsection{Localisation of Global Models}

Some recent works in the literature introduce models with both global and local parameters, with the most relevant works in this space being DeepGLO \citep{sen_2019_think} and the winning approach of the M4 forecasting competition, Exponential Smoothing-Recurrent Neural Network \citep[ES-RNN,][]{ref_1}. However, we focus in our work on methods that can be generically applied to existing GFM models, to enable them to achieve more localised learning.

ES-RNN not only has local and global parameters, but it also has a mechanism for localisation, called the ensemble of specialists. This architecture first assigns the series randomly to train a set of GFM experts. Then, it identifies the most suitable series to train each specialist based on the errors corresponding to a validation dataset. This procedure is iteratively repeated until the average forecasting error on the validation set starts growing. Finally, forecasts for each time series are computed by aggregating the forecasts of GFM specialists that correspond to each time series. In this approach, data heterogeneity is addressed by selecting the most suitable series to train each specialist, and generating the final forecasts using multiple GFM specialists. 

The only prior work that uses time series clustering for localising and addressing data heterogeneity of GFMs, to the best of our knowledge, is the work of \citet{ref_2}. This work implements feature-based clustering approaches using K-means \citep{ref_90}, DBScan \citep{Ester1996ADA}, Partition Around Medoids \citep[PAM,][]{kaufman_1990_pam} and Snob \citep{wallace1994intrinsic} as base clustering algorithms. For each identified cluster, those authors build a separate GFM, using RNN as the base forecasting algorithm. That work also has been extended to address real-world forecasting challenges in retail and healthcare \citep{ref_13,Bandara2020-en}. However, these approaches have the limitation of information loss that may occur due to the suboptimal grouping of time series. As the method makes a hard binary decision about which GFM to use for a particular time series, based on the clustering, a wrong selection here leads to a sub-optimal model being used. Our proposed framework addresses this limitation by aggregating the forecasts generated by a set of clusters in multiple iterations. On top of this, we use both feature-based and distance-based clustering techniques to identify subgroups of time series, where prior work has been only based on feature-based clustering.

\citet{pablo_2020_principles} see clustering as a method to add more complexity to a GFM. Those authors show that increasing model complexity can improve the forecasting accuracy, thus localising GFMs by using techniques such as clustering supports to improve the performance of GFMs. They only consider random clustering for their experiments. We use their work as a theoretical base and a starting point, also addressing data heterogeneity by using more sophisticated clustering techniques, and therewith also increasing model complexity in a directed and focused way.

The main notion of GFMs is typically that all series used for training are relevant for evaluation, which is why it is beneficial to develop a GFM that will perform well on average across these series (and localise accordingly). Another related topic is data augmentation, where, e.g., \citet{BANDARA2021108148} use simulated series to improve the performance of GFMs. However, as here the GFM does not need to perform well on the augmented series, here the GFM needs to be trained in a pooled way across real and augmented data, without localisation, so that the GFM can generalise across both real and augmented data, to prevent overfitting on the real data.

\subsection{Ensembling for Forecasting: Forecast Combinations}

In time series forecasting, ensembling is known under the name of forecast combination~\citep{ref_58, ref_59, ref_62}. As in many ensembling techniques, forecast combination techniques usually train multiple submodels independently, and then aggregate over the submodels to compute final forecasts. The submodel aggregation technique plays a crucial role in forecast combination approaches. Simple averaging and weighted averaging are  widely used to aggregate forecasts \citep{ref_58}. 

\citet{krogh_1995_neural} show that the squared error of a linearly combined ensemble model can be explained as the addition of two distinct components: average squared errors of the individual base models and a term that quantifies the interactions or diversity among the base model predictions. Those authors show that this second term which represents the diversity is always positive. Thus, the squared error of the ensemble model is always less than or equal to the average squared error of the individual base models. This phenomenon shows that higher diversity of base models produces a larger gain for the ensemble model over the average performance of the individual base models. Furthermore, \citet{brown_2005_diversity} show that diversity acts as an extra degree of freedom in the bias-variance trade-off which allows the ensemble model to find functions that are hard to find with an individual forecasting model. 
Thus, ensemble models are expected to provide better results when submodel forecasts are uncorrelated, and heterogeneous base learners are more likely to generate uncorrelated forecasts \citep{ref_51,ref_52}. \citet{ref_53} propose an ensemble of bagged trees approach, where diversity among the model trees is increased by using different time series representations. \citet{ref_51} propose an ensemble approach with dynamic heterogeneous submodels that uses a weighting scheme. These heterogeneous ensemble models have been used to address many real-world forecasting problems, such as power forecasting \citep{ref_55}, electricity load forecasting \citep{ref_56}, and price forecasting \citep{ref_57}. 

Ensembling and localisation have been used together to improve the prediction accuracy. \citet{masoudnia2014mixture} discuss prediction methods that incorporate a mixture of experts concept where multiple neural networks are trained for different parts of the problem space. The work by \citet{laurinec2019density} provides insights on how clustering based ensembling techniques can be used specifically to forecast electricity consumption.  

Meta-learning based weighted ensembles are also popular. The second-winning approach in the M4 forecasting competition, Feature-based Forecast Model Averaging  \citep[FFORMA,][]{ref_3} uses a weighted ensembling approach that combines base models using a meta-learner, which is trained using time series features. The third-winning approach \citep{PAWLIKOWSKI202093} also uses an ensemble technique to optimally combine the forecasts generated by a set of statistical models. The meta-learning approach proposed by \citet{cerqueira2019arbitrage} uses the prediction of the loss of sub-models to determine the weights that should be used when combining their forecasts.

\subsection{Time Series Clustering}

Clustering, or identifying similar groups of data that have similar characteristics, is a widely used approach to handle data heterogeneity \citep{ref_45}. Three main classes of time series clustering approaches are distance-based, feature-based, and model-based approaches \citep{ref_44}. The distance-based clustering techniques group time series based on the similarity of raw observations of time series. Here, the similarity of point values is captured using a distance metric, such as Euclidean distance, Manhattan distance, Pearson correlation, Spearman correlation, or DTW. As the performance of distance-based clustering techniques depends heavily on the distance metric used, choosing the appropriate distance measure can be a challenging task in this approach. In contrast, feature-based clustering techniques first extract a set of features from each time series. Then, a traditional clustering technique is applied to the extracted feature matrix to identify similar groups of time series. Compared to distance-based clustering, feature-based clustering approaches are typically more interpretable and more resilient to missing data \citep{ref_46}. On the other hand, model-based clustering assumes the data are originally generated from a group of models and therefore attempts to recover the models from the data \citep{ref_47}. Here, expectation-maximisation algorithms are used to assign each series to the cluster with maximum likelihood.

\section{Methods}
\label{sec:methods}

This section first gives a brief overview of the proposed methodology and the primary prediction models used to implement the GFMs in this study and their corresponding preprocessing techniques. Then, we describe the GFM-based ensemble variants introduced.

\subsection{Methodology Overview: Localisation of GFMs Through Ensembling}
\label{sec:problem_statement}

\begin{figure}[htb]
    \centering
    \includegraphics[width=\textwidth]{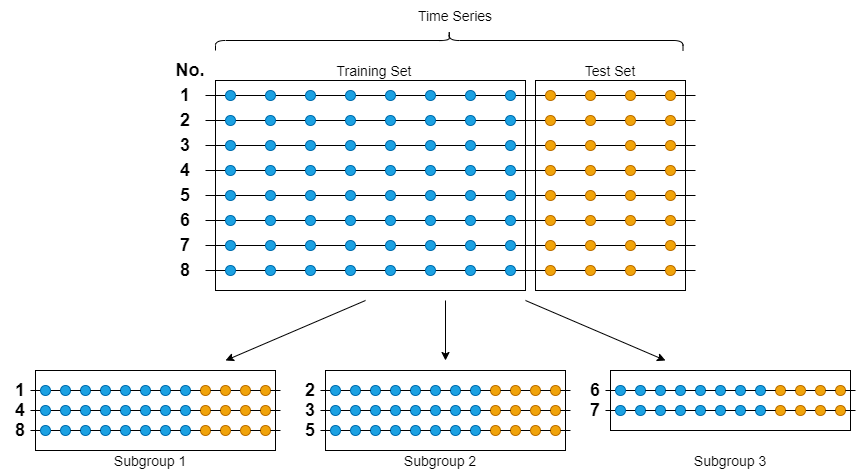}
    \caption{Visualisation of the training and test sets in GFM context. Series are clustered into different groups based on their similarity and a localised GFM is trained per each series cluster.}
    \label{fig:problem_statement}
\end{figure}

Time series forecasting refers to predicting the future values of a given set of time series based on their past values. In our experimental framework, we consider a dataset as a set of time series. As shown in Figure \ref{fig:problem_statement}, we split each time series in a dataset into training and test parts where the training parts are used to train GFMs and the test parts represent the actual values corresponding with the expected forecast horizon. In the GFM context, the training set consists of the past values from a group of time series, whereas the test set consists of their corresponding future values.

Our proposed framework first clusters a given collection of time series based on their similarity, and then trains a set of localised GFMs, one per each time series cluster, where the training set of each GFM consists of the training parts of the corresponding time series cluster. The future values of each time series are obtained using their corresponding localised GFM. This procedure is repeated multiple times either by changing the number of clusters or cluster seed. Therefore, for each series, we obtain multiple forecasts provided by a set of localised GFMs in multiple iterations. The final forecasts for a given series are obtained by averaging the forecasts provided by the corresponding localised GFMs. Though there is research in the literature around the use of weighted averages and other techniques to obtain the final forecasts \citep{ref_62, ref_51}, and there are situations where weighted averaging provides superior results over simple averaging \citep{ref_3, godahewa_2020_weekly}, a consensus in the forecasting literature is that a simple average is hard to beat and often performs best \citep{ref_58, genre_2013_average}. Thus, we limit our experiments to simple averages as we deem this particular question not to be the main focus of our paper.

\subsection{Global Forecasting Models}

We use FFNNs, RNNs, and PR models as the baseline GFMs in our work. FFNNs have the universal function approximation property and therewith are capable of approximating any (non-linear) function \citep{ref_71}. RNNs belong to the NN family and they are especially suitable for sequence modelling problems \citep{ref_8} due to their capability of addressing temporal order and temporal dependencies of sequences. Furthermore, global RNNs have been recently used as a central part of the winning method of the M4 forecasting competition, making them one of the state-of-the-art models in the forecasting space \citep{ref_1}. As both FFNNs and RNNs are non-linear GFMs, we also consider linear GFMs in our work. Hence, we also consider PR models \citep{ref_114} in our work which use linear functions of lagged values of time series to predict their future. Further details of these GFMs are detailed in the following.    

\subsubsection{Feed-Forward Neural Networks}

FFNN is a strong non-linear standard model, which has the universal function approximation property \citep{ref_71}. 
For further details of FFNNs, we refer to \citet{Goodfellow-et-al-2016}. In our work we use the implementation available in R in the \verb|nnet| function from the \verb|nnet| package \citep{ref_110}.
To use an FFNN for time series forecasting, the time series are preprocessed as explained in Section \ref{sec:gfm_preprocessing}. We then train the FFNNs using the single-step ahead forecasting strategy, where models are iteratively used to obtain the forecasts for the required forecast horizon~\citep{taieb2012review}.

\subsubsection{Recurrent Neural Networks with Long Short-Term Memory Cells}
\label{sec:rnn_lstm}

RNNs are a type of NNs that are especially suited for sequence modelling \citep{ref_8}. In RNNs,  feedback loops address the temporal order and temporal dependencies of sequences, which may be less aptly handled by traditional NNs \citep{ref_7}. Recently, RNN-based GFMs are becoming a popular NN architecture among forecasting practitioners \citep{ref_1, ref_2, ref_6, ref_13, ref_24, ref_99}, and have been used as the primary architecture in the winning solution of the M4 forecasting competition.
Based on recommendations by \citet{ref_6}, we use the LSTM network as our primary RNN architecture and implement the Stacking Layers design pattern to train the network \citep{ref_12, godahewa_2020_weekly}. However, we highlight that the proposed methodology can be generalised to any RNN variant such as Elman RNN \citep{ref_8}, Gated Recurrent Units \citep[GRU,][]{ref_10}, and others. We implement the proposed RNN architectures using \verb|TensorFlow| \citep{ref_26}, based on the framework developed by \citet{ref_6}.
Similar to FFNN, time series are preprocessed as discussed in Section \ref{sec:gfm_preprocessing}. Then, we use the RNN stacked architecture with the multi-step ahead forecasting strategy to train the network, following the procedure recommended by \citet{ref_6} .

\subsubsection{Pooled Regression Models}
PR models \citep{ref_114} are, similar to FFNNs and in contrast to RNNs, purely auto-regressive, and use a linear function of the lagged values of a time series to predict its future. They estimate a set of constant coefficients corresponding to each considered lagged value. PR models train on a pool of time series during model training and the coefficients are computed by minimising the overall training error. The lagged values are then multiplied by their corresponding coefficients and an intercept is added to obtain the final forecasts. PR models use the single-step ahead forecasting strategy and therefore the models are iteratively used to obtain the forecasts for the required forecast horizon.
In our experiments, we use the \verb|glm| function in the \verb|glmnet| package \citep{ref_111} to implement the proposed PR variants.

\subsubsection{Preprocessing for Global Forecasting Models}
\label{sec:gfm_preprocessing}
\begin{figure}
    \centering
    \includegraphics[width=\textwidth]{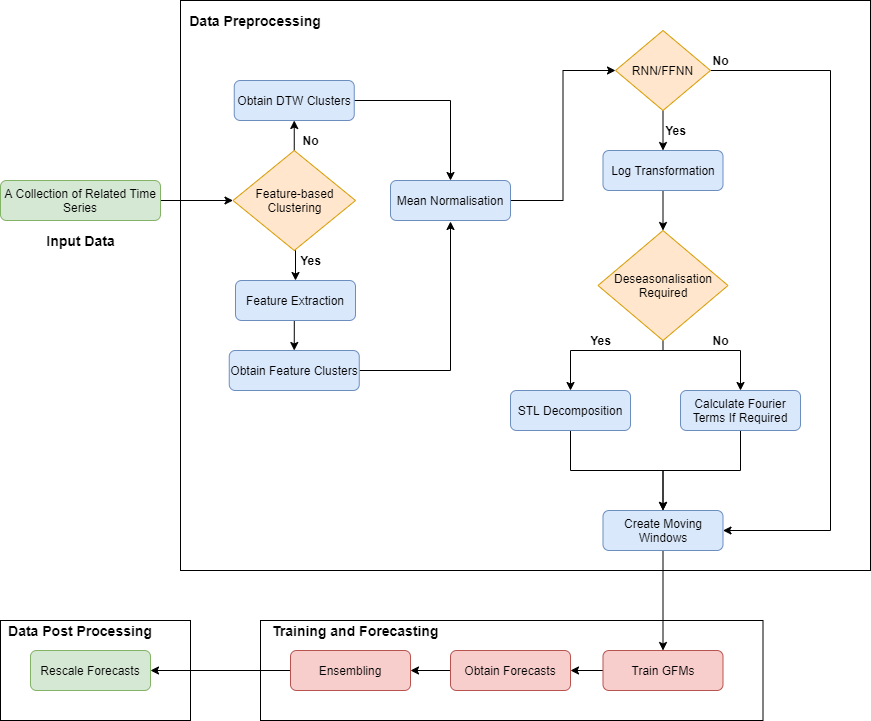}
    \caption{Overall workflow of the proposed ensembled GFMs. The data are first preprocessed based on the GFM type. The preprocessed data are then used to train multiple GFMs whereas their forecasts are ensembled accordingly and rescaled to obtain the final forecasts.}
    \label{fig:workflow}
\end{figure}

For FFNN and RNN models, the time series are first normalised by dividing them by their corresponding mean values. Next, the normalised series are transformed to a log scale to stabilise the variance and to avoid placing the outputs in the saturated areas of the NN's activation function.
There are two main approaches to handle the seasonality of time series when using RNNs: deseasonalisation or using original observations with seasonal exogenous variables \citep{ref_24}. Depending on the characteristics of the datasets and the results of prior work \citep{ref_6}, we determine the datasets that require deseasonalisation (Section \ref{sec:datasets}).  If required, deseasonalisation is performed by applying the Seasonal and Trend Decomposition using Loess (STL) method \citep{ref_21}, and  the deseasonalised series are used to train the RNN model. For the datasets that do not require deseasonalisation, the preprocessed series are directly used to train the RNN model, along with the seasonal regressors. In particular, for multiseasonal datasets, we use Fourier terms \citep{ref_4}, a set of sine and cosine terms, together with the preprocessed series when training the RNN model following the procedure suggested by \citet{ref_24}.  The \verb|stl| and \verb|fourier| functions in the \verb|forecast| package \citep{ref_22} are used, respectively, for deseasonalisation and Fourier terms calculation. 
Finally, the moving window scheme is applied to the preprocessed series to create the input and output window pairs required to train the NN models \citep{ref_6}.
On the other hand, when implementing the PR models, only mean normalisation is used to preprocess the time series. This is because, unlike in NNs, PR models do not have activation functions, and therewith do not show saturation effects in the way NNs do~\citep{ref_6}.
Once the forecasts are obtained from the GFMs, to bring them to the original scale, the corresponding preprocessing techniques are reversely applied. Figure \ref{fig:workflow} shows the overall workflow of the proposed ensembled GFMs.

\subsection{Ensemble Models to Localise Global Forecasting Models}
\label{sec:ensemble}
In the following, we describe the different types of GFM-based ensemble models used in our experiments.

\subsubsection{Ensemble of Specialists/Experts}
\label{sec:ensemble_of_experts}

The ensemble of specialists was proposed by \citet{ref_1}, and used by the same author in ES-RNN, the winning solution of the M4 forecasting competition \citep{ref_30}. The ES-RNN implementation uses RNN with LSTM cells as the primary specialist/expert prediction unit.

\begin{algorithm}
\fontsize{10}{10}
\caption{Ensemble of Specialists Algorithm}
\label{alg1}

\begin{algorithmic}[1]

\State \COMMENT{\textbf{ Parameters \\
training\_set - Contains the training parts of each series of the dataset\\
validation\_set - Contains the training and validation parts of each series of the dataset\\
test\_set - Contains all full-length time series \\
validation\_results - Contains actual data corresponding with the validation parts of each series\\
num\_of\_specialists - Number of GFMs to be used in the algorithm\\
N - Number of top GFMs whose forecasts are combined to obtain the final forecasts 
}}\\

\Procedure{ensemble\_of\_specialists}{$training\_set$, $validation\_set$, $test\_set$, $validation\_results$, $num\_of\_specialists$, $N$} 

\State $gfms$ $\gets$ $[]$ \COMMENT{\textbf{Stores GFM specialists}}
\State $train\_series$ $\gets$ $[]$ \COMMENT{\textbf{Stores training series corresponding with each GFM}}
\State $val\_predictions$ $\gets$ $[]$ \COMMENT{\textbf{Stores validation forecasts given by GFMs for a particular iteration}}
\State $test\_predictions$ $\gets$ $[]$ \COMMENT{\textbf{Stores GFM forecasts corresponding with testing period}}
\State $final\_predictions$ $\gets$ $[]$ \COMMENT{\textbf{Stores final output forecasts}}
\State $val\_errors$ $\gets$ $[]$ \COMMENT{\textbf{Stores validation errors given by each GFM in a particular iteration}}
\State $prev\_errors$ $\gets$ $[]$ \COMMENT{\textbf{Stores validation errors given by GFMs in the previous iteration}}
\State $val\_error\_growing$ $\gets$ $FALSE$ \COMMENT{\textbf{Whether the validation set error is growing or not}}
\State $iterations$ $\gets$ 1

\For{$i$ in 1 to $num\_of\_specialists$}
\State	$gfms[i]$ $\gets$ $init\_gfm\_model()$
\State  $train\_series[i]$ $\gets$ $random(training\_set, 0.5)$ \COMMENT{\textbf{Randomly allocate training series for GFMs}}
\EndFor

\While{$val\_error\_growing$ $==$ $FALSE$}

	\If{$iterations$ $>$ 1} 
	    \State \COMMENT{\textbf{Re-assign series to their corresponding top performing GFMs}}	
		\State $train\_series$ $\gets$ $reassign\_series(val\_errors, N)$ 
	\EndIf

	\For{$i$ in 1 to $num\_of\_specialists$}
		\State  $gfm[i]$ $\gets$ $train\_gfm(gfm[i], train\_series[i])$ \COMMENT{\textbf{Train GFMs}} 
		 
		\State  $val\_predictions[i]$ $\gets$ $test\_gfm(gfm[i], validation\_set)$ \COMMENT{\textbf{Get validation forecasts}}
        \State \COMMENT{\textbf{ Calculate validation errors}}
		\State $val\_errors[i]$ $\gets$ $calc\_errors(val\_predictions[i], validation\_results)$
	\EndFor
	
	\If{$iterations$ $>$ 1} 
	\State \COMMENT{\textbf{Check whether the average validation error is growing or not}}
		\State $val\_error\_growing$ $\gets$ $check\_error\_growing(val\_errors, prev\_errors)$
	\EndIf 

	\State $prev\_errors$ $\gets$ $val\_errors$
	\State $iterations$ $\gets$ $iterations + 1$
\EndWhile
\algstore{myalg}
\end{algorithmic}
\end{algorithm}

\begin{algorithm}  
\fontsize{10}{10}                   
\begin{algorithmic} [1]                   
\algrestore{myalg}

\State \COMMENT{\textbf{Get forecasts corresponding with the testing period from trained GFMs}}
\For{$i$ in 1 to $specialists$}
	\State  $test\_predictions[i]$ $\gets$ $test\_gfm(gfm[i], test\_set)$
\EndFor

\State \COMMENT{\textbf{The final forecasts of each series are obtained by averaging the forecasts provided by its top performing GFMs}}
\For{$s$ in 1 to len($test\_set)$}
	\State $final\_predictions[s]$ $\gets$ $combine\_best\_predictions(test\_predictions, N, s)$
\EndFor

\Return $final\_predictions$
\EndProcedure
\end{algorithmic}
\end{algorithm}

Algorithm \ref{alg1} shows the training and testing phases of the ensemble of specialists method. In the training phase, the algorithm learns a group of GFMs simultaneously to obtain the best GFMs to model each time series. In the first iteration, 50\% of the time series are randomly allocated for each GFM to train. Then, each GFM is trained using its assigned series and the forecasts are obtained from the trained GFMs for a validation dataset. The GFMs are then ranked for each time series based on the validation errors and the time series are re-allocated for their corresponding top $N$ (in our case $N=2$) GFMs. The series allocation procedure is repeated for each GFM until the average validation error starts to grow. In the testing phase, for each test series, the average of the forecasts corresponding to the top $N$ GFMs forms the final forecast. 
We tune the number of specialists ($K$) as a hyperparameter using a validation dataset (refer to Section \ref{sec:hyperparameters}).

\subsubsection{Clustering-based Ensemble Models}
\label{sec:feature_clustering}

In this approach, we use a combination of feature and distance-based clustering techniques to build ensemble models. Table \ref{tab:clustering_techniques_overview} 
summarises the clustering techniques used in our experiments. Here, the Category column indicates the approach to which the clustering algorithm is applied, and the Package and Function columns provide a reference to the software implementation used in our experiments.

\begin{table*}[htb]
\centering
\begin{tabular}{llll}
\hline
\textbf{Clustering Technique} & \textbf{Category} & \textbf{Package}
& \textbf{Function}\\
\hline
K-means &  feature-based & stats \citep{stat_2020_pkg} & kmeans \\
K-means++ &  feature-based & LICORS \citep{goerg_2013_licors} & kmeanspp \\
K-medoids with DTW Distance & distance-based &  dtwclust \citep{ref_96} & tsclust \\
\hline
\end{tabular}
\caption{Overview of Clustering Techniques}
\label{tab:clustering_techniques_overview}
\end{table*}

For the feature-based clustering approach, we use K-means \citep{ref_90} and K-means++ \citep{ref_91} as the primary clustering algorithms. Here, the K-means++ algorithm is used to address the weak centroid initialisation problem present in the traditional K-means algorithm.
We extract a set of self-explainable features \citep{ref_88} from each time series and cluster them to identify similar groups of time series. This approach is similar to the methodology suggested by \citet{ref_2}. The extracted features include mean, variance, first order of autocorrelation, trend, linearity, curvature, entropy, lumpiness, spikiness, level shift, variance change, flat spots, crossing points, maximum Kullback-Leibler (KL) shift and time KL shift. We use the \verb|tsfeatures| function from the \verb|tsfeatures| package \citep{ref_89} to extract these time series features. 
In this way, we represent each time series as a vector of features. We then apply the target clustering algorithms to determine the optimal feature grouping, which represents the subgroups of time series that are used to train the GFMs. 
There have been other feature extraction procedures proposed in the literature, such as catch22 \citep{lubba_2019_catch22} or tsfresh \citep{CHRIST201872}, and their use would most likely result in different clustering outcomes. However, \verb|tsfeatures| is a popular and well-established package in the forecasting space, and has been used in a similar setting by \citet{ref_2}, which is why we focus on this package.

To implement the distance-based clustering approach, we use the K-medoids clustering \citep{ref_95} algorithm, using DTW as the primary distance measure. The K-medoids algorithm handles noise and outliers better compared with K-means. As distance-based clustering algorithms work directly on the raw observations of the time series, a prior preprocessing phase is not necessary.

Based on the above clustering approaches, we introduce the following model variants to our framework:

\begin{description}
    \item[GFM.Cluster.Number] We apply clustering algorithms multiple times by varying the number of clusters. Multiple GFMs are then trained per each identified cluster, for multiple iterations, generating multiple forecasts per each series. These forecasts are then averaged to obtain the final forecasts. 
    \item[GFM.Cluster.Seed] We use the elbow method \citep{ref_45} to determine the optimal number of clusters, and then apply the K-means or K-means++ algorithms multiple times with the optimal number of clusters by changing the cluster seed. Finally, multiple GFMs are trained per each cluster, for multiple iterations and the forecasts are ensembled following the same procedure used in GFM.Cluster.Number.
    
To determine the optimal number of clusters, the elbow method runs a clustering algorithm multiple times by gradually increasing the number of clusters. It calculates the sum of squared errors in clustering for each considered number of clusters and plots the errors against their respective number of clusters. At the beginning, the error is expected to quickly decrease as the number of clusters increases. At some point, the error reduction rate slows down leading to a situation of diminishing returns. This point can be determined as the point of maximum curvature (forming an elbow) in the plot and the number of clusters corresponding with that point is considered as the optimal number of clusters to be used with the considered clustering algorithm. We visualise the results of the elbow method with our experimental datasets using the R package \verb|factoextra| \citep{ref_97}.
\end{description}

\begin{figure}[htb]
    \centering
    \includegraphics[width=\textwidth]{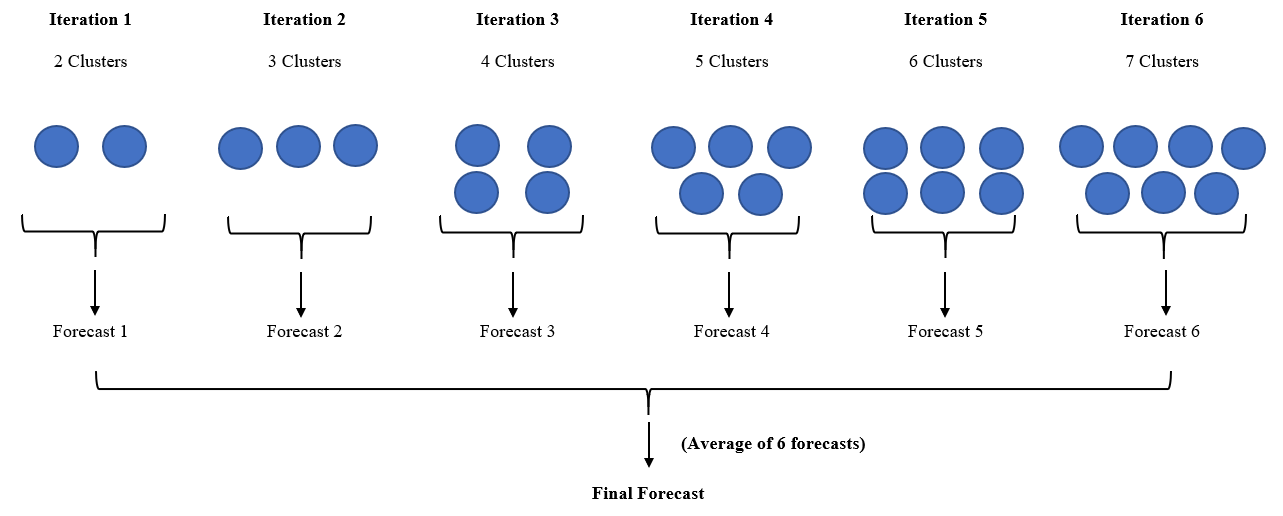}
    \caption{In the GFM.Cluster.Number variant, we consider the cluster range from two to seven. Six forecasts are generated for each time series by varying the cluster number. Finally, these forecasts are averaged to obtain the final forecasts. On the other hand, the GFM.Cluster.Seed variant uses the same number of clusters, the optimal cluster number, varying the clusters from iteration to iteration by varying the cluster seeds.}
    \label{fig:cluster_ensembles}
\end{figure}

Figure \ref{fig:cluster_ensembles} illustrates the GFM.Cluster.Number variant used in our experiments, and Algorithms \ref{alg2} and \ref{alg3} show the training and testing phases of the GFM.Cluster.Number and GFM.Cluster.Seed models, respectively. We note that the range of cluster numbers for GFM.Cluster.Number is defined in a way that the optimal number of clusters found by the elbow method is placed within that range.

\sloppy The feature clustering technique is applied for both the  GFM.Cluster.Number and GFM.Cluster.Seed variants, whereas the distance-based clustering technique is only applied for the GFM.Cluster.Number variant. This is because, the GFM.Cluster.Seed algorithm requires an optimal number of clusters to operate. According to the best of our knowledge, there is no straightforward method to identify the optimal number of clusters for K-medoids clustering with DTW distances. So, altogether there are five clustered ensemble models that we use in the experiments: K-means, K-means++ and K-medoids (DTW) with the GFM.Cluster.Number variant, and K-means and K-means++ with the GFM.Cluster.Seed variant.

\begin{algorithm}
\fontsize{10}{10}
\begin{algorithmic}[1]

\State \COMMENT{\textbf{ Parameters \\
training\_set - Contains the training parts of each series of the dataset\\
test\_set - Contains all full-length time series \\
cluster\_range - A vector of clusters numbers \\
clustering\_method - The clustering technique e.g. K-means
}}

\\

\Procedure{ensemble\_of\_clustering\_GFM.Cluster.Number}{$training\_set$, $test\_set$, $cluster\_range$, $clustering\_method$} 

\State $predictions$ $\gets$ $[][]$ \COMMENT{\textbf{Stores forecasts of all iterations}}
\State $final\_predictions$ $\gets$ $[]$ \COMMENT{\textbf{Stores final average forecasts}} 

\State $iterations$ $\gets$ 1
\State $seed$ $\gets$ $generate\_seed()$ 

\For{$cluster\_num$ in $cluster\_range$}
  \State \COMMENT{\textbf{Cluster all series for the numbers specified in cluster\_range iteratively}}
  \State  $cluster\_ids$ $\gets$ $cluster\_series(test\_set, cluster\_num, clustering\_method, seed)$ 
  \For{$c$ in $cluster\_num$}
    \State $series\_ids$ $\gets$ $cluster\_ids[c]$
    \State  $gfm$ $\gets$ $init\_gfm\_model()$
  	\State  $gfm$ $\gets$ $train\_gfm(gfm, train\_series[series\_ids])$ \COMMENT{\textbf{Train GFMs using clustered series}}
  	 \State \COMMENT{\textbf{Obtain forecasts for the test series in clusters using the trained GFMs}}
  	\State  $predictions[series\_ids][iterations]$ $\gets$ $test\_gfm(gfm, test\_series[series\_ids])$
  \EndFor 
  \State $iterations$ $\gets$ $iterations$ + 1
\EndFor

\State \COMMENT{\textbf{Calculate the average of forecasts produced in each iteration}}
\State $final\_predictions$ $\gets$ $row\_wise\_avg(predictions)$
	
\Return $final\_predictions$
\EndProcedure
\caption{GFM.Cluster.Number Algorithm}
\label{alg2}
\end{algorithmic}
\end{algorithm}

\begin{algorithm}
\fontsize{10}{10}
\begin{algorithmic}[1]

\State \COMMENT{\textbf{ Parameters \\
training\_set - Contains the training parts of each series of the dataset\\
test\_set - Contains all full-length time series \\
optimal\_num\_of\_clusters - The optimal number of clusters to be used given by the elbow method \\
num\_of\_iterations - Number of cluster generating iterations \\
clustering\_method - The clustering technique e.g. K-means
}}

\\

\Procedure{ensemble\_of\_clustering\_GFM.Cluster.Seed}{$training\_set$, $test\_set$, $optimal\_num\_of\_clusters$, $num\_of\_iterations$, $clustering\_method$} 

\State $predictions$ $\gets$ $[][]$ \COMMENT{\textbf{Stores forecasts of all iterations}}
\State $final\_predictions$ $\gets$ $[]$ \COMMENT{\textbf{Stores final average forecasts}}

\For{$iter$ in $num\_of\_iteration$}
  \State $seed$ $\gets$ $generate\_seed()$ \COMMENT{\textbf{Generate a random seed in each iteration}}
  
  \State \COMMENT{\textbf{Cluster all series into the optimal number of clusters with the generated seed}}
  \State  $cluster\_ids$ $\gets$ $cluster\_series(test\_set, optimal\_num\_of\_clusters, clustering\_method, seed)$ 
  
  \For{$c$ in $optimal\_num\_of\_clusters$}
    \State $series\_ids$ $\gets$ $cluster\_ids[c]$
    \State  $gfm$ $\gets$ $init\_gfm\_model()$
  	\State  $gfm$ $\gets$ $train\_gfm(gfm, train\_series[series\_ids])$ \COMMENT{\textbf{Train GFMs using clustered series}}
  	\State \COMMENT{\textbf{Obtain forecasts for the test series in clusters using the trained GFMs}}
  	\State  $predictions[series\_ids][iter]$ $\gets$ $test\_gfm(gfm, test\_series[series\_ids])$
  \EndFor 
\EndFor

\State \COMMENT{\textbf{Calculate the average of forecasts produced in each iteration}}
\State $final\_predictions$ $\gets$ $row\_wise\_avg(predictions)$
	
\Return $final\_predictions$
\EndProcedure
\caption{GFM.Cluster.Seed Algorithm}
\label{alg3}
\end{algorithmic}
\end{algorithm}

\subsubsection{Forecast Combinations of Global and Local Forecasting Models}
\label{sec:average_clustering_ets}

We combine the predictions of six RNN-based GFMs with the predictions of two traditional univariate forecasting models. The RNN-based models are a single RNN and five clustered ensemble models based on RNNs, namely: RNN.Cluster.Number with K-means, K-means++ and K-medoids, and RNN.Cluster.Seed with K-means and K-means++ methods. The traditional techniques are ETS \citep{ref_112} and ARIMA \citep{ref_113}. The RNN-based forecasts are averaged with ETS forecasts as well as with both ETS and ARIMA forecasts, making 12 forecast combination models.

When applying forecast combinations to multi-seasonal benchmark datasets, we replace the ETS and ARIMA methods with the Trigonometric, Box-Cox, ARMA, Trend, Seasonal \citep[TBATS,][]{ref_93} and the Dynamic Harmonic Regression ARIMA \citep[DHR-ARIMA,][]{ref_94} methods, which are capable of forecasting time series with multiple seasonal patterns.

\subsubsection{Ensemble Baseline}
\label{sec:seed_rnn}

Ensembling can be used in many different ways with different objectives. E.g., boosting lowers bias, and bagging lowers variance, by addressing model, parameter, and/or data uncertainty. Therefore, we employ an ensembling baseline to be able to quantify how much accuracy is gained from localisation versus from other effects. In particular, we implement a straightforward approach to address the parameter uncertainty of GFMs by training GFMs multiple times using different seeds. We train global RNN/FFNN models using all the available time series with five different seeds. The forecasts generated by these five seeds are then averaged to obtain the final forecasts.

\section{Experimental Framework}
\label{sec:framework}

In this section, we discuss the benchmark datasets, hyper-parameter optimisation approaches, error metrics, and benchmarks used in our experiments.

\subsection{Datasets}
\label{sec:datasets}

We use eight publicly available datasets\footnote{The experimental datasets are available at \url{https://drive.google.com/drive/folders/16xqLEFyLn_gJcXrIp1LWyAD_KvJjB5Hn?usp=sharing}.} to evaluate the performance of our proposed variants. Table \ref{tab:dataset_overview} provides summary statistics of these benchmark datasets, and a brief overview is as follows: 

\begin{itemize}
    \item M3 Dataset \citep{ref_5}: Monthly dataset from the M3 forecasting competition. This dataset contains six subcategories of time series, namely micro, macro, industry, demo, finance, and other. When executing our proposed variants, we use this information as the domain knowledge and a separate global model is trained for each subcategory. 
    
    \item M4 Monthly Dataset \citep{ref_30}: Monthly dataset from the M4 forecasting competition. This dataset contains the same six subcategories as the M3 dataset, and the GFMs are built per subcategory accordingly.
    
    \item CIF 2016 Dataset \citep{ref_31}: Monthly dataset from the CIF 2016 forecasting competition, containing 24 time series from the banking domain and 48 simulated series. We use all simulated and real-world series to train models and to evaluate their performance with this dataset. Hence, the simulated series also contribute to the final forecasting accuracy of the dataset.
    
    \item Kaggle Wikipedia Web Traffic Dataset \citep{ref_32}: A subset of time series from the Kaggle Wikipedia Web Traffic forecasting competition that contains numbers of daily hits/traffic for a given set of Wikipedia web pages. The missing observations of this dataset are treated as zeros, following the procedure suggested by \citet{ref_6}.
        
	\item Ausgrid Half-Hourly Dataset \citep{ref_33}: A half-hourly time series dataset, representing the energy consumption of Australian households. In this dataset, time series may exhibit multiple seasonal patterns, such as hourly, daily, and weekly seasonality.

	\item Ausgrid Monthly Dataset \citep{ref_33}: A collection of monthly time series, representing the energy consumption of Australian households. In our experiments, we only include time series with more than 49 observations, to avoid possible constraints in our hyperparameter tuning process. 

	\item Tourism Quarterly Dataset \citep{Athanasopoulos_2011_tourism}: Quarterly dataset from a tourism forecasting competition. 

	\item Hospital Dataset: A collection of monthly time series, representing patient counts related to medical products. This dataset was extracted from the R package \verb|expsmooth| \citep{hyndman_2015_expsmooth}. 
\end{itemize}

\begin{table*}
\centering
\resizebox{1.0\textwidth}{!}{
\begin{tabular}{lccccc}
\hline
\textbf{Dataset Name} & \textbf{No. of} & \textbf{Forecast} & \textbf{Frequency} & \textbf{Minimum} & \textbf{Maximum} \\
 & \textbf{Time Series} & \textbf{Horizon} &  & \textbf{Length} & \textbf{Length} \\
\hline
M3 &  1428 & 18 & Monthly & 48 & 126\\
M4 &  48000 & 18 & Monthly & 42 & 2794\\
CIF 2016 & 72 & 6, 12 & Monthly & 22 & 108\\
Wikipedia Web Traffic & 997 & 59 & Daily & 550 & 550\\
Ausgrid Half-Hourly & 300 & 96 & Half-Hourly & 1488 & 1488\\
Ausgrid Monthly & 1405 & 12 & Monthly & 38 & 88\\
Tourism & 427 & 8 & Quarterly & 22 & 122 \\
Hospital & 767 & 12 & Monthly & 72 & 72 \\
\hline
\end{tabular}}
\caption{Datasets Information}
\label{tab:dataset_overview}
\end{table*}

\subsection{Dataset Specific Preprocessing}

The input window size of NNs or the number of lagged values used in the PR models are determined using the heuristic suggested by \citet{ref_6}.  
Furthermore, we deseasonalise the M3, M4, CIF 2016 and Kaggle web traffic time series datasets to train RNN models, following the results of~\citet{ref_6}. 
For the Ausgrid monthly, Ausgrid half-hourly dataset, Tourism and Hospital datasets, we run preliminary experiments with different methods to address the seasonality, and as a result deseasonalise the Tourism and Hospital series, do not deseasonalise the Ausgrid monthly series, and use Fourier terms to handle multiple seasonalities in the Ausgrid half-hourly dataset.

\subsection{Hyperparameter Tuning}
\label{sec:hyperparameters}

\begin{table*}
\begin{center}
\resizebox{1.0\textwidth}{!}{
\begin{tabular}{lccccccccccc}
\hline
Dataset & \multicolumn{8}{c}{RNN} & \multicolumn{2}{c}{FFNN} & \multicolumn{1}{c}{Common}\\
\cline{2-9}
\cline{10-12}
& Batch Size & Epochs & Epoch Size & Std. Noise & L2 Reg. & Cell Dim. & Layers & Std. Initializer & Nodes & Decay & Specialists\\
\hline
M3(Mic) & 40 - 80 & 2 - 30 & 2 - 10 & 0.0001 - 0.0008 & 0.0001 - 0.0008 & 20 - 50 & 1 - 2 & 0.0001 - 0.0008 & 1 - 12 & 0 - 0.1 & 2 - 7\\
M3(Mac) & 30 - 50 & 2 - 30 & 2 - 10 & 0.0001 - 0.0008 & 0.0001 - 0.0008 & 20 - 50 & 1 - 2 & 0.0001 - 0.0008 & 1 - 12 & 0 - 0.1 & 2 - 7\\
M3(Ind) & 30 - 50 & 2 - 30 & 2 - 10 & 0.0001 - 0.0008 & 0.0001 - 0.0008 & 20 - 50 & 1 - 2 & 0.0001 - 0.0008 & 1 - 12 & 0 - 0.1 & 2 - 7\\
M3(Dem) & 10 - 20 & 2 - 30 & 2 - 10 & 0.0001 - 0.0008 & 0.0001 - 0.0008 & 20 - 50 & 1 - 2 & 0.0001 - 0.0008 & 1 - 12 & 0 - 0.1 & 2 - 7\\
M3(Fin) & 10 - 25 & 2 - 30 & 2 - 10 & 0.0001 - 0.0008 & 0.0001 - 0.0008 & 20 - 50 & 1 - 2 & 0.0001 - 0.0008 & 1 - 12 & 0 - 0.1 & 2 - 7\\
M3(Oth) & 5 - 10 & 2 - 30 & 5 - 20 & 0.0001 - 0.0008 & 0.0001 - 0.0008 & 20 - 50 & 1 - 2 & 0.0001 - 0.0008 & 1 - 12 & 0 - 0.1 & 2 - 7\\
M4(Mic) & 1000 - 1500 & 2 - 25 & 2 - 10 & 0.0001 - 0.0008 & 0.0001 - 0.0008 & 20 - 50 & 1 - 2 & 0.0001 - 0.0008 & 1 - 12 & 0 - 0.1 & 2 - 7\\
M4(Mac) & 1000 - 1500 & 2 - 25 & 2 - 10 & 0.0001 - 0.0008 & 0.0001 - 0.0008 & 20 - 50 & 1 - 2 & 0.0001 - 0.0008 & 1 - 12 & 0 - 0.1 & 2 - 7\\
M4(Ind) & 1000 - 1500 & 2 - 25 & 2 - 10 & 0.0001 - 0.0008 & 0.0001 - 0.0008 & 20 - 50 & 1 - 2 & 0.0001 - 0.0008 & 1 - 12 & 0 - 0.1 & 2 - 7\\
M4(Dem) & 850 - 950 & 2 - 25 & 2 - 10 & 0.0001 - 0.0008 & 0.0001 - 0.0008 & 20 - 50 & 1 - 2 & 0.0001 - 0.0008 & 1 - 12 & 0 - 0.1 & 2 - 7\\
M4(Fin) & 1000 - 1500 & 2 - 25 & 2 - 10 & 0.0001 - 0.0008 & 0.0001 - 0.0008 & 20 - 50 & 1 - 2 & 0.0001 - 0.0008 & 1 - 12 & 0 - 0.1 & 2 - 7\\
M4(Oth) & 30 - 40 & 2 - 25 & 2 - 10 & 0.0001 - 0.0008 & 0.0001 - 0.0008 & 20 - 25 & 1 - 2 & 0.0001 - 0.0008 & 1 - 12 & 0 - 0.1 & 2 - 7\\
CIF (12) & 5 - 10 & 2 - 25 & 5 - 20 & 0.01 - 0.08 & 0.0001 - 0.0008 & 20 - 50 & 1 - 2 & 0.0001 - 0.0008 & 1 - 12 & 0 - 0.1 & 2 - 7\\
CIF (6) & 2 - 5 & 2 - 30 & 5 - 15 & 0.0001 - 0.0008 & 0.0001 - 0.0008 & 20 - 50 & 1 - 5 & 0.0001 - 0.0008 & 1 - 12 & 0 - 0.1 & 2 - 7\\
Wikipedia & 100 - 150 & 2 - 25 & 2 - 10 & 0.0001 - 0.0008 & 0.0001 - 0.0008 & 20 - 25 & 1 - 2 & 0.0001 - 0.0008 & 1 - 12 & 0 - 0.1 & 2 - 7\\
Ausgrid (HH) & 20 - 50 & 10 - 40 & 2 - 10 & 0.0001 - 0.0008 & 0.0001 - 0.0008 & 20 - 50 & 1 - 2 & 0.0001 - 0.0008 & 1 - 12 & 0 - 0.1 & 2 - 7\\
Ausgrid (M) & 20 - 100 & 2 - 25 & 2 - 10 & 0.0001 - 0.0008 & 0.0001 - 0.0008 & 20 - 50 & 1 - 2 & 0.0001 - 0.0008 & 1 - 12 & 0 - 0.1 & 2 - 7\\
Tourism & 20 - 100 & 2 - 25 & 2 - 10 & 0.0001 - 0.0008 & 0.0001 - 0.0008 & 20 - 50 & 1 - 2 & 0.0001 - 0.0008 & 1 - 12 & 0 - 0.1 & 2 - 7\\
Hospital & 20 - 100 & 2 - 25 & 2 - 10 & 0.0001 - 0.0008 & 0.0001 - 0.0008 & 20 - 50 & 1 - 2 & 0.0001 - 0.0008 & 1 - 12 & 0 - 0.1 & 2 - 7
\\\hline
\end{tabular}
}
\caption{Initial Hyperparameter Ranges}
\label{tab:hyperparameter_ranges}
\end{center}
\end{table*}

The base learners RNN and FFNN have various hyper-parameters. The RNN models include number of epochs, epoch size, mini-batch size, cell dimension, number of hidden layers, L2 regularisation weights, standard deviation of random normal initialiser and standard deviation of Gaussian noise \citep{ref_6}. The COntinuous COin Betting \citep[COCOB, ][]{ref_29} algorithm is used as the learning algorithm to train the RNN-based variants, which does not require tuning of the learning rate.
The FFNN models have number of nodes in the hidden layers and learning rate decay as model hyper-parameters. Furthermore, the number of specialists required by the ensemble of specialists model is also tuned as a separate hyper-parameter for each base learner. Meanwhile, the PR-based model variants do not have hyper-parameters that require tuning.
We autonomously determine the optimal values of the hyper-parameters of our base models using the sequential model-based algorithm configuration (SMAC), a variant of Bayesian Optimisation proposed by \citet{ref_34}. In our experiments, the Python implementation of SMAC \citep{ref_35} is used. The initial hyperparameter ranges used in the SMAC algorithm for each dataset are shown in Table \ref{tab:hyperparameter_ranges}. The SMAC algorithm then finds the optimal values for each hyperparameter within the defined range using a predefined number of iterations or a time limit.
We use a sequence of observations from the end of each series of length equivalent to the intended forecast horizon as the validation set for hyperparameter tuning. The GFMs are trained using the remaining observations of each series and the forecasts corresponding to the validation set are used to optimise the hyperparameters.

In general, the hyperparameter optimisation process is conducted using all the time series in a dataset. However, our ensemble models use different groups of time series to train multiple submodels and hence, performing separate hyperparameter tuning for each submodel is not efficient. 
To overcome this issue, a set of time series is randomly chosen from a given dataset with a size of the number of series divided by the maximum number of submodels considered in one iteration of an ensemble model, e.g., the maximum number of submodels used in one iteration of the ensemble of specialists or clustered ensemble models. The hyperparameters are tuned based on this chosen set of time series. The same set of hyperparameters are then used to train each submodel in the considered ensemble models. We also train non-ensemble models using the same set of hyperparameters, for a consistent comparison against ensemble models.

\subsection{Error Metrics}
\label{sec:error_metrics}
We measure the performance of our models using the symmetric Mean Absolute Percentage Error (sMAPE), Mean Absolute Scaled Error \citep[MASE,][]{ref_36}, Mean Absolute Error \citep[MAE,][]{ref_95} and Root Mean Squared Error (RMSE) which are commonly used error metrics in the field of time series forecasting. Equations \ref{eqn:smape}, \ref{eqn:mase}, \ref{eqn:mae} and \ref{eqn:rmse} define the sMAPE, MASE, MAE and RMSE error metrics, respectively. Here, $M$ is the number of instances in the training set, $N$ is the number of data points in the test set, $F_k$ are the forecasts, $Y_k$ are the actual values for the required forecast horizon and $S$ is the length of the seasonal cycle. 

\begin{equation}
\label{eqn:smape}
    sMAPE = \frac{100\%}{N}\sum_{k=1}^{N} \frac{|F_{k} - Y_{k}|}{(|Y_{k}| + |F_{k}|)/2} 
\end{equation}

\begin{equation}
\label{eqn:mase}
    MASE = \frac{\sum_{k=1}^{N} |F_{k} - Y_{k}|}{\frac{N}{M - S}\sum_{k=S+1}^{M} |Y_{k} - Y_{k - S}|}
\end{equation}

\begin{equation}
\label{eqn:mae}
    MAE = \frac{\sum_{k=1}^{N}{|F_{k} - Y_{k}|}}{N}
\end{equation}

\begin{equation}
\label{eqn:rmse}
    RMSE = \sqrt{\frac{\sum_{k=1}^{N}{|F_{k} - Y_{k}|}^2}{N}}
\end{equation}

For datasets containing zeros, namely the Ausgrid datasets and the Kaggle web traffic dataset, we use a variant of the sMAPE error measure proposed by \citet{ref_14}, to avoid possible division by zero problems. To achieve this, the denominator of Equation \ref{eqn:smape} is changed to $max(|Y_{k}| + |F_{k}| + \epsilon, 0.5 + \epsilon)$, where we set $\epsilon$ to its proposed default value of 0.1.

Since all these error measures are defined for each time series, we calculate the mean and median values of them across a dataset to measure the model performance. Therefore, each model is evaluated using eight error metrics: mean sMAPE, median sMAPE, mean MASE, median MASE, mean MAE, median MAE, mean RMSE and median RMSE.

\subsection{Benchmarks and Variants}
\label{sec:benchmarks}

To compare against our proposed ensemble variants, we use two univariate forecasting models, ETS \citep{ref_112} and ARIMA \citep{ref_113}, which are competitive and commonly used benchmarks in time series forecasting research. Furthermore, we use TBATS \citep{ref_93} and DHR-ARIMA \citep{ref_94} as the univariate benchmarks for the Ausgrid half-hourly dataset, as ETS and ARIMA are not well-suited to handle the multiple seasonal patterns present in half-hourly data. 
We use the default implementations available from the \verb|forecast| package of these methods. Furthermore, we use a grid-search approach to determine the optimal number of Fourier terms for DHR-ARIMA, which controls the smoothness of the seasonal pattern to be modelled.

We also use 2 deep learning benchmarks, DeepAR \citep{ref_99} and N-BEATS \citep{oreshkin_2019_nbeats}. DeepAR is a global auto-regressive RNN used with probabilistic forecasting. N-BEATS is an interpretable deep neural network based forecasting framework which uses backward and forward residual links with a deep stack of fully-connected layers. In our work, we use the implementations of DeepAR and N-BEATS available from the Python package \verb|GluonTS| \citep{gluonts_2020_alexandrov} with their default hyperparameters. 

As a benchmark based on gradient boosted trees, we use CatBoost \citep{NEURIPS2018_14491b75}. In contrast to other popular implementations of gradient boosted trees, it considers the order of the data points during model training, thus making it more appropriate for time series forecasting. In our work, we use the implementation of CatBoost available from the R package \verb|catboost| \citep{NEURIPS2018_14491b75} with its default hyperparameters that are expected to provide good results.

As an ensemble benchmark, we use FFORMA, which combines the forecasts from a set of univariate models based on the weights obtained by training a gradient boosted tree model using 42 features, which are calculated from the original time series. This method achieved the overall second place in the M4 forecasting competition \citep{ref_3}. In our work, we use the implementation of FFORMA available from the \verb|M4metalearning| package \citep{m4metalearning_2020_pablo}.
Apart from these, the following benchmarks are used in our experiments. 

\begin{description}
    \item[Baseline] The forecasts provided by a single FFNN, RNN, or PR model trained across a full dataset.
    \item[Kmeans.OC] Use K-means clustering with optimal number of clusters to cluster the features, then train separate GFMs per cluster, so that forecasts for a series are obtained by one particular GFM.
    \item[KmeansPlus.OC] The same as Kmeans.OC, but using the K-means++ clustering algorithm.
    \item[Xmeans] The same as the last two methods, but with Xmeans clustering \citep{ref_92}, which is able to automatically identify an optimal number of clusters. We use the \verb|XMeans| function in the \verb|RWeka| package \citep{ref_98} to implement Xmeans clustering. 
\end{description}   

Furthermore, we benchmark the proposed methods against random clustering methods, as suggestsed in \citet{pablo_2020_principles}. Random clustering methods are proposed as a way to increase model complexity. In particular, we use three models based on random clustering as benchmarks, as follows:

\begin{description}
 	\item[Random.OC] Cluster series randomly, train separate GFMs with each series cluster and obtain forecasts for each series once with its corresponding GFM.
    \item[Random.Number] Cluster series randomly and build  the GFM.Cluster.Number model. 
    \item[Random.Seed] Cluster series randomly and build the GFM.Cluster.Seed model.
\end{description}

The five variants of the clustered ensemble models (Section \ref{sec:feature_clustering}), ensemble of specialists and ensemble of NNs are used as follows:

\begin{description}
    \item[DTW.Number] Apply K-medoids clustering with DTW distances for series and build \newline GFM.Cluster.Number model.
    \item[Kmeans.Number] Apply K-means clustering for features and build the GFM.Cluster.Number model.
    \item[KmeansPlus.Number] Apply K-means++ clustering for features and build the GFM.Cluster.Number model.
    \item[Kmeans.Seed] Apply K-means clustering for features and build the GFM.Cluster.Seed model.
    \item[KmeansPlus.Seed] Apply K-means++ clustering for features and build the GFM.Cluster.Seed model.
    \item[Ensemble.Specialists] Ensemble of Specialists using RNNs, FFNNs and PR as the base models.
    \item[Ensemble.Seed] Train NNs with different seeds in multiple iterations and average the results.
\end{description}

\subsection{Statistical Testing of the Results}
\label{sec:statistical_tests}

We consider the recommendations given by \citet{demsar_2006_statistical} to analyse the statistical significance of results provided by different forecasting methods across the experimental datasets. The null hypothesis of no difference between the methods is tested by applying a Friedman test \citep{friedman_1940_comparison}, and if the null hypothesis is rejected, a pairwise statistical
comparison: the Wilcoxon signed-rank test with Holm’s alpha
correction ($\alpha = 0.05$) is used following the recommendations given by \citet{benavoli_2016_ranks}. The results of statistical tests are visualised using a critical difference diagram \citep{demsar_2006_statistical}.
We consider ETS and TBATS, and ARIMA and DHR-ARIMA as pairs together for statistical testing as we do not evaluate these models with each dataset. All other 45 models including FFORMA, DeepAR, N-BEATS, CatBoost and GFMs based on RNNs, FFNNs, and PR are individually considered for statistical testing. We do not consider the M4 monthly dataset for statistical testing, as due to its much large size compared with the other datasets it would dominate the results. Finally, the mean sMAPE results of 47 models across the other 7 experimental datasets are used for statistical testing.
We use the Python implementation \verb|cd-diagram| available on Github to plot the critical difference diagram \citep{IsmailFawaz2018deep}.

\section{Analysis of Results}
\label{sec:results}

This section provides a comprehensive analysis of the results of all considered models across the experimental datasets.
Table \ref{tab:sMAPE_all} summarises the overall performance of the proposed variants and the benchmarks, in terms of mean sMAPE across all experimental datasets. The conclusions we draw by analysing the mean sMAPE values are in agreement with the other seven error metrics that are available in the Online Appendix\footnote{The online appendix is available at \url{https://drive.google.com/file/d/1lJuGNvGA8O-U0OSh8KHUEPeNvhESv75W/view?usp=sharing}.}.

\begin{table}
\begin{center}
\resizebox{\textwidth}{!}{
\begin{tabular}{lcccccccc}
  \hline
Model & M3 & M4 & CIF & Ausgrid HH & Ausgrid M & Kaggle & Tourism & Hospital \\ 
  \hline
  ETS & \textbf{14.14} & 13.53 & 12.18 & - & 26.49 &  52.54 & 15.07 & 17.46 \\ 
  ARIMA & 14.25 & 13.08 & \textbf{11.70} & - & 26.69 &  \textbf{47.96} & 16.58 & 17.79\\
  TBATS & - & - & - & \textbf{36.20} & - & - & - & -\\
  DHR-ARIMA & - & - & - & 39.31 & - & - & - & - \\
  \hline 
  FFORMA & 14.51 & \textbf{12.64} & 13.23 & 46.40 & 26.30 & 48.65 & 14.88 & \textbf{17.13}\\
  CatBoost & 16.41 & 14.05 & 14.87 & 44.07 & 24.92 & 51.50 & 16.53 & 18.09 \\ 
  DeepAR & 15.74 & 14.29 & 13.58 & 68.47 & \textbf{24.75} & 57.79 & 15.29 & 17.45 \\ 
  N-BEATS & 14.76 & 13.40 & 11.72 & 44.89 & 45.08 & 51.82 & \textbf{14.45} & 17.77 \\ 
  \hline
  \hline
      
  RNN Baseline & 14.82 & 14.21 & 11.47 & 34.96 & 26.76 &  46.52 & \textbf{15.99} & 19.80\\ 
  RNN Kmeans.OC & 14.49 & 13.90 & 11.14 & 35.27 & 27.37 & 47.49 & 16.32 & 19.45\\ 
  RNN KmeansPlus.OC & 14.56 & 14.18 & 11.17 & 35.54 & 25.64 & 47.39 & 16.28 & 19.48\\ 
  RNN Xmeans & 14.27 & 13.65 & 10.69 & 35.27 & 25.69 & 47.69 & 16.44 & 19.50\\ 
  \hline 
  RNN DTW.Number & \textbf{14.14} &  \textbf{13.48} & 11.10 & \textbf{34.79} & 25.64 & 46.99 & 16.46 & 19.40\\ 
  RNN Kmeans.Number  & 14.33 & 13.65 & 10.77 & 34.83 & 25.23 & 46.88 & 16.39 & \textbf{19.30}\\ 
  RNN KmeansPlus.Number  & 14.32 & 13.64 & 10.92 & 34.88 & \textbf{25.04} & 47.35 & 16.37 & 19.35\\ 
  RNN Kmeans.Seed & 14.40 & 14.05 & 11.14 & 34.82 & 26.92 & 47.05 & 16.31 & 19.39\\ 
  RNN KmeansPlus.Seed & 14.48 & 13.99 & 11.17 & 34.91 & 25.78 & 46.88 & 16.26 & 19.44\\ 
  RNN Ensemble.Specialists & 14.55 & 13.96 & \textbf{10.55} & 34.92 & 25.56 & 46.63 & 16.25 & 19.50\\
  RNN Ensemble.Seed & 14.74 & 14.39 & 10.98 & 34.94 & 26.72 &  \textbf{46.42} & 16.21 & 19.76\\  
  \hline
  RNN Random.OC & 14.78 & 14.79 & 10.91 & 36.91 & 25.58 & 46.91 & 16.50 & 19.34\\ 
  RNN Random.Number & 14.64 & 14.04 & 10.81 & 35.28 & 25.69 & 47.03 & 16.49 & 19.41\\ 
  RNN Random.Seed & 14.59 & 14.82 & 10.84 & 35.49 & 25.25 & 47.14 & 16.24 & 19.34\\ 
  \hline
  \hline
    
  FFNN Baseline & 15.60 & 15.75 & 16.30 & 38.72 & 25.03 &  47.69 & 16.49 & 18.76\\ 
  FFNN Kmeans.OC & 15.71 & 14.96 & 13.54 & 45.04 & 25.79 & 48.00 
  & 18.30 & 18.10\\ 
  FFNN KmeansPlus.OC & 15.19 & 14.49 & 13.29 & 41.80 & 25.32 & 47.76 & 17.86 & 18.48\\ 
  FFNN Xmeans & 15.26 & 14.28 & 14.29 & 40.03 & 25.00 & 47.55 & 18.57 & 18.15\\ 
  \hline 
  FFNN DTW.Number & \textbf{14.89} &  \textbf{13.75} & 13.44 & 39.79 & \textbf{24.97} & 47.75 & 16.79 & \textbf{17.61}\\ 
  FFNN Kmeans.Number  & 15.18 & 14.15 & 14.06 & 37.91 & 25.38 & 47.57 & 17.36 & 17.92\\ 
  FFNN KmeansPlus.Number  & 15.02 & 14.15 & 13.80 & 37.87 & 25.38 & \textbf{47.52} & 17.64 & 17.98\\ 
  FFNN Kmeans.Seed & 15.57 & 14.50 & \textbf{13.11} & 38.19 & 25.39 & 47.62 & 18.06 & 17.90
  \\ 
  FFNN KmeansPlus.Seed & 15.37 & 14.50 & 13.52 & 38.21 & 25.37 & 47.56 & 17.63 & 18.07\\ 
  FFNN Ensemble.Specialists & 15.61 & 14.46 & 15.03 & 39.34 & 24.99 & 47.80 & \textbf{15.70} & 18.12\\ 
  FFNN Ensemble.Seed & 15.36 & 14.88 & 15.06 & \textbf{37.36} & 25.00 &  47.81 & 16.49 & 18.36\\ 
  \hline
  FFNN Random.OC & 15.79 & 14.95 & 15.69 & 38.51 & 25.24 & 47.72 
  & 16.77 & 18.63\\ 
  FFNN Random.Number & 15.42 & 14.55 & 14.82 & 39.14 & 25.15 & 47.82 & 16.51 & 18.38\\ 
  FFNN Random.Seed & 15.39 & 14.62 & 14.40 & 37.62 & 25.21 & 47.88 & 16.42 & 18.27
  \\ 
  \hline
  \hline
    
  PR Baseline & 14.36 &  13.77 & 12.86 & 39.81 & 24.08 & 131.90 & 15.86 & 17.56\\ 
  PR Kmeans.OC & 14.29 & 13.30 & 12.24 & 39.74 & 24.24 &  90.00 & 15.34 & 17.82\\ 
  PR KmeansPlus.OC & 14.22 & 23.12 & \textbf{12.19} & 39.82 & 24.27 &  90.00 & 15.43 & 17.90\\ 
  PR Xmeans & 14.31 & 13.04 & 12.79 & 39.72 & \textbf{24.04} &  90.01 & 15.45 & 17.89\\ 
  \hline
  PR DTW.Number & 14.04 &  \textbf{12.87}  & 12.49 & 39.56 & 24.20 & 103.37 & 15.59 & 17.87\\ 
  PR Kmeans.Number & \textbf{13.91} & 13.00 & 12.33 & 39.71 & 24.06 &  91.48 & 15.16 & \textbf{17.43}\\ 
  PR KmeansPlus.Number & 14.03 & 13.00 & 12.60 & 39.96 & 24.09 &  \textbf{88.73} & 15.17 & 17.56\\ 
  PR Kmeans.Seed & 14.09 & 13.29 & 12.21 & 39.75 & 24.24 &  90.00 & 15.34 & 17.87\\ 
  PR KmeansPlus.Seed & 14.22 & 13.30 & 12.24 & 39.83 & 24.27 &  90.00 & 15.34 & 17.90\\ 
  PR Ensemble.Specialists & 14.09 & 13.63 & 12.47 & \textbf{39.47} & 24.28 & 116.09 & \textbf{15.06} & 17.54\\ 
  \hline
  PR Random.OC & 14.41 & 13.77 & 12.86 & 39.86 & 24.06 & 131.38  & 15.85 & 17.62\\ 
  PR Random.Number & 14.27 & 13.73 & 12.64 & 39.80 & 24.07 & 128.56 & 15.79 & 17.59\\ 
  PR Random.Seed & 14.24 & 13.76 & 12.67 & 39.81 & 24.09 & 127.96 & 15.84 & 17.58 \\ \hline
\end{tabular}
}
\caption{Mean sMAPE results. The benchmark models, the RNN-based models, the FFNN-based models, and the PR-based models are separated with double lines. The corresponding best-performing model for each section is highlighted in boldface.}
\label{tab:sMAPE_all}
\end{center}
\end{table}

\subsection{Relative Performance of Clustered Ensemble Models}
\label{sec:ensemble_clustering_performance}
\begin{figure}[htb]
\centering
  \includegraphics[width=\textwidth]{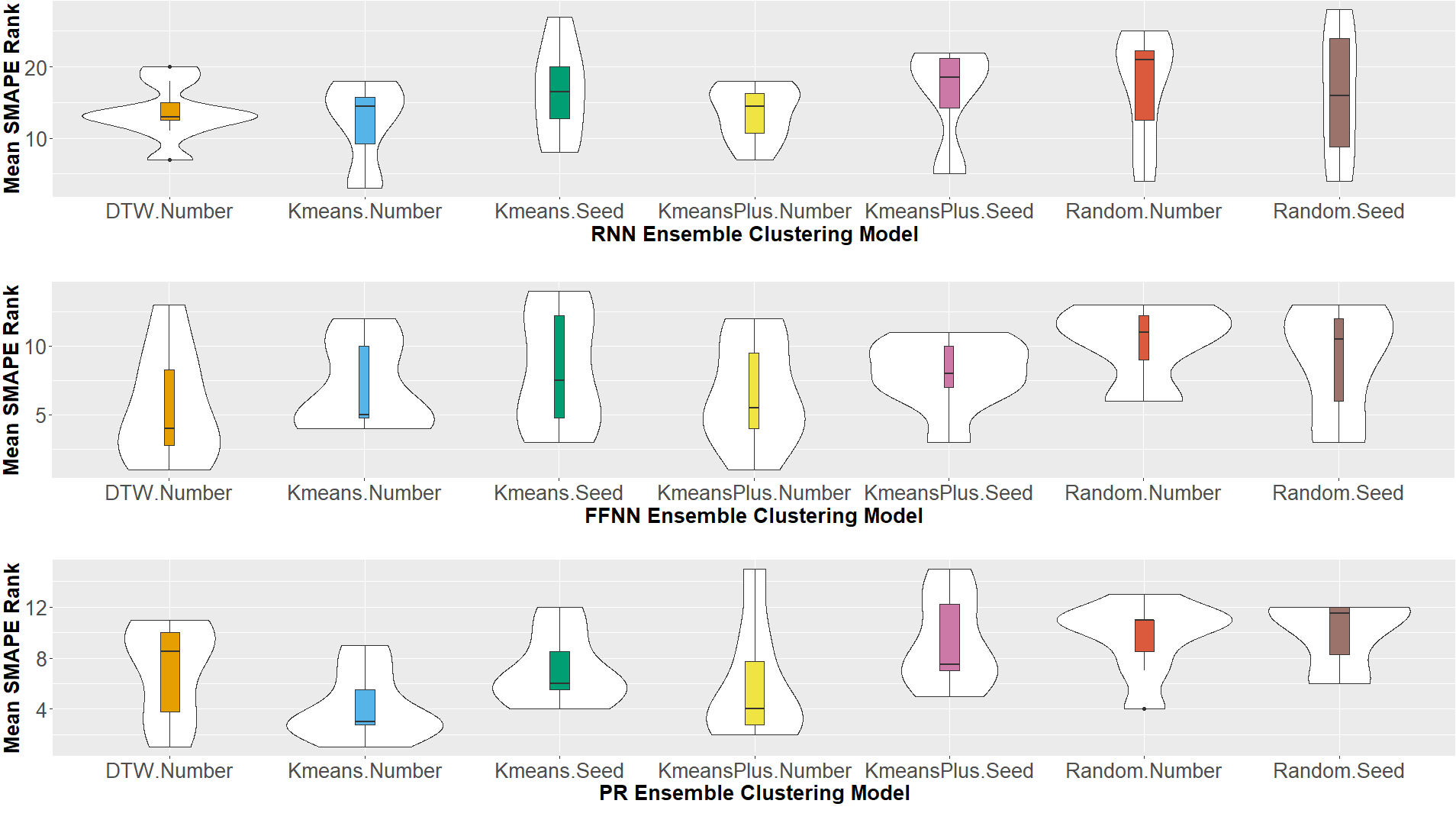}
  \caption{Relative Performance of Clustered Ensemble Models}~\label{fig:ensemble_clustering_models}
\end{figure}

We compare the relative performance of seven clustered ensemble models: Kmeans.Number, KmeansPlus.Number, Random.Number, DTW.Number, Kmeans.Seed, KmeansPlus.Seed and Random.Seed across all datasets. Figure \ref{fig:ensemble_clustering_models} shows the mean sMAPE ranks of these clustered ensemble models implemented using the 3 base GFMs: RNNs, FFNNs and PR models using violin plots enabling to compare their performance in terms of the inter-quantile ranges and density distributions of error ranking.

The clustered ensemble models that correspond to different GFMs show varied performances. According to Figure \ref{fig:ensemble_clustering_models}, we see that for the RNN and FFNN models, the DTW.Number cluster variant obtains the best mean sMAPE rank whereas the Kmeans.Number cluster variant obtains the best mean sMAPE rank for the PR models. The inter-quantile ranges of DTW.Number and Kmeans.Number cluster variants are respectively lower across NN and PR models. Furthermore, higher density areas of the DTW.Number and Kmeans.Number cluster variants are respectively located much lower along the y axis across NN and PR models.  The results also indicate that overall, the GFM.Cluster.Number based cluster variants outperform the GFM.Cluster.Seed based cluster variants.

With respect to the random clustering models, overall, Random.Number and Random.Seed perform worst since their inter-quantile ranges and density distributions are located much higher along the y-axis compared to other variants. Hence, our experiments clearly show that increasing the model complexity by using sophisticated clustering techniques is able to improve the forecasting accuracy.

\subsection{Relative Performance of Clustered Non-Ensemble Models}

\begin{figure}[htb]
\centering
  \includegraphics[width=\textwidth]{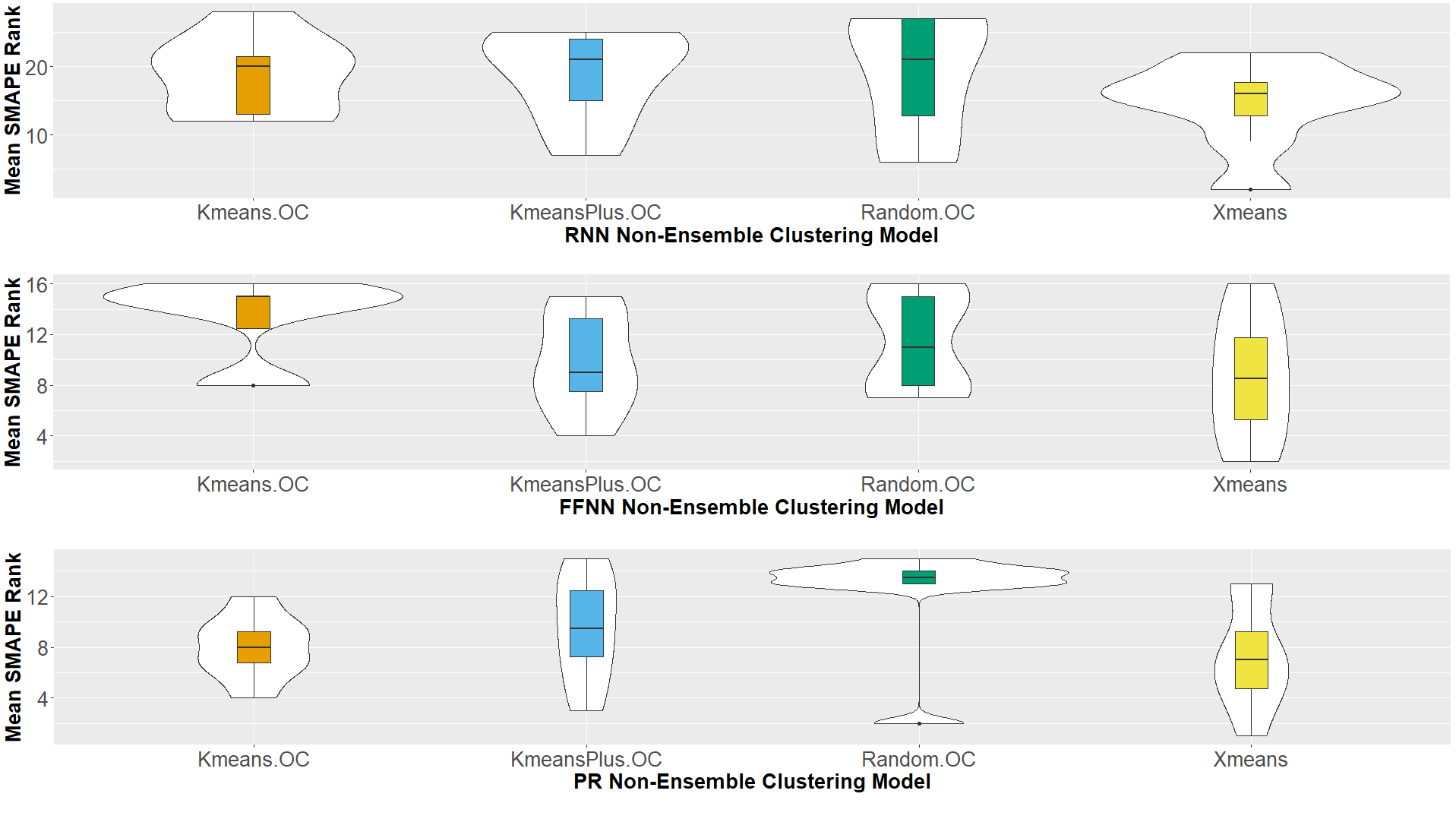}
  \caption{Relative Performance of Clustered Non-Ensemble Models}~\label{fig:non_ensemble_clustering_models}
\end{figure}

Figure \ref{fig:non_ensemble_clustering_models} shows the relative performance of clustered non-ensemble models: Kmeans.OC, KmeansPlus.OC, Random.OC and Xmeans based on mean sMAPE ranks for the 3 base GFMs. Xmeans shows the best performance over all GFMs. Random.OC shows the worst performance with RNNs and PR models, whereas Kmeans.OC shows the worst performance with FFNNs since the higher density areas of those variants are located much higher across the y axis.

Xmeans identifies the optimal number of clusters to be used by itself and therefore, it is capable of reducing the prediction error compared to Kmeans.OC and KmeansPlus.OC where the optimal number of clusters is determined using the elbow method \citep{ref_45} with a graphical approach.

\subsection{Performance of Clustered Ensemble Models Vs. Clustered Non-Ensemble Models}
In Table \ref{tab:sMAPE_all} we see that our proposed five clustered ensemble variants: DTW.Number, Kmeans.Number, KmeansPlus.Number, Kmeans.Seed, and KmeansPlus.Seed show a better performance compared to clustered non-ensemble models across all the datasets with all GFMs except for 3 cases: RNN Xmeans and PR KmeansPlus.OC on the CIF 2016 dataset, PR Xmeans on the Ausgrid monthly dataset.

In non-ensemble clustering, information loss can happen due to training multiple models with different subgroups of datasets. Ensembling of the results of clustering models in multiple iterations helps to mitigate the information loss while properly addressing data heterogeneity during model training. With this, the methods can provide more accurate forecasts than the clustered non-ensemble methods.

\subsection{Performance of Non-Clustering based Ensemble Models and Deep Learning Benchmarks}
\label{sec:non_clustering_ensembles}
From Table \ref{tab:sMAPE_all}, we can furthermore see that the Ensemble.Specialists variant shows an overall better performance compared to the Baseline models across all GFMs. However, our feature and distance-based clustered ensemble models outperform the Ensemble.Specialists for all GFMs across all benchmark datasets, except for 6 cases: CIF 2016, Kaggle web traffic and Tourism on RNNs, Tourism on FFNNs and Ausgrid half-hourly and Tourism on PR. Although the ensemble of specialists considers data heterogeneity during model training, it does not outperform the clustered ensemble techniques in many cases. For our experiments, we use a generic version of Ensemble.Specialists. We also note that the original implementation of the Ensemble.Specialists method uses a different base method than we do. Thus, the Ensemble.Specialists variant may require advanced preprocessing techniques, advanced GFM architectures based on datasets and a set of hyperparameter values chosen based on expert knowledge to provide better results, where in our version, the design choices do not depend on the dataset or expert knowledge. 

Overall, the Ensemble.Seed variants show a better performance compared to the Baseline models. The RNN Baseline only outperforms the RNN Ensemble.Seed on the M4 and Tourism datasets and the FFNN Baseline only outperforms the FFNN Ensemble.Seed variant on the Kaggle web traffic dataset. 
As the Ensemble.Seed models address parameter uncertainty they are able to achieve better performance than the Baseline models. 
However, all other ensemble variants show an overall better performance compared with the Ensemble.Seed models.

The FFORMA method outperforms our clustered ensemble models on the M4, Tourism and Hospital datasets across all GFMs, and shows better performance compared to FFNN and PR on the M3 dataset and the Kaggle web traffic dataset, respectively.
However, overall our clustered ensemble models outperform FFORMA. Moreover, Ensemble.Specialists models show a mixed performance compared to FFORMA across all baseline GFMs. The poor performance of the FFORMA method on the Kaggle web traffic and Ausgrid half-hourly datasets can be attributed to its instability around forecasting time series with consecutive zeros/constant values. Also, the limited amount of time series available for model training may be a reason for the underperformance of FFORMA on the CIF 2016 dataset.

Our clustered ensemble models outperform the considered gradient boosted trees based benchmark, CatBoost, for all GFMs across all datasets, except for 5 cases: Ausgrid monthly and Hospital on RNNs, Ausgrid monthly and Tourism on FFNNs and Kaggle web traffic on PR.

Furthermore, our clustered ensemble models overall outperform DeepAR across all datasets and N-BEATS across 7 datasets: M3, M4, CIF 2016, Kaggle web traffic, Ausgrid half hourly, Ausgrid monthly and Hospital.

\subsection{Performance of Global Models and Traditional Univariate Forecasting Models}
\label{sec:local_global_performance}

According to the results of Table \ref{tab:sMAPE_all}, overall, RNNs show a better performance than FFNNs across all experimental datasets. Furthermore, RNNs and PR-based GFM variants show a competitive performance. RNNs outperform the PR models for the CIF 2016, Kaggle web traffic, and Ausgrid half-hourly datasets, while for the remaining 5 datasets, PR models demonstrate a better performance than RNNs.

We further compare the performance of the traditional univariate forecasting models and the three GFMs: RNNs, FFNNs, and PR models. As shown in Table \ref{tab:sMAPE_all}, even the RNN Baseline outperforms the individual univariate forecasting models on the CIF 2016, Ausgrid half-hourly and Kaggle web traffic datasets. The PR Baseline model outperforms the univariate forecasting models on the Ausgrid monthly dataset. For the Tourism dataset, all GFM baselines and for the Hospital dataset, the PR baseline show a better performance than ARIMA. For the M3 and M4 datasets, the traditional univariate benchmarks show a better performance than the Baseline GFMs. But our clustered ensemble models and ensemble of specialist models based on GFMs, outperform the univariate forecasting benchmarks across all benchmark datasets.

\begin{table*}[h]
\begin{center}
\resizebox{\textwidth}{!}{
\begin{tabular}{llcccccccc}
  \hline
Local Models & RNN Variant & M3 & M4 & CIF & Ausgrid HH & Ausgrid M & Kaggle & Tourism & Hospital\\ 
  \hline
 & Baseline & 13.86 & 13.10 & 11.32 & 34.30 & 24.18 & 48.49 & 14.39 & 17.79\\ 
 & DTW.Number & 13.67 &  12.80 & 11.18 & 34.27 & 23.93 & 49.13 & 14.50 & 17.68\\ 
 ETS/TBATS & Kmeans.Number & 13.68 & 12.95 & \textbf{11.03} & 34.22 & 23.83 & 49.10 & 14.55 & 17.65\\
 & KmeansPlus.Number & 13.66 & 12.95 & 11.07 & 34.23 & 23.83 & 49.37 & 14.56 & 17.67\\ 
 & Kmeans.Seed & 13.71 & 13.02 & 11.17 & 34.18 & 24.14 & 49.23 & 14.54 & 17.70\\ 
 & KmeansPlus.Seed & 13.73 & 13.03 & 11.20 & \textbf{34.12} & 24.10 & 49.10 & 14.51 & 17.71\\ 
 
 \hline 
 
 & Baseline & 13.54 & 12.73 & 11.25 & 34.75 & 23.75 & \textbf{48.07} & \textbf{14.26} & 17.36\\
 & DTW.Number & 13.45 & \textbf{12.57} & 11.21 & 34.73 & 23.70 & 48.49 & 14.35 & 17.32\\ 
 ETS/TBATS, & Kmeans.Number & 13.46 & 12.67 & 11.10 & 34.70 & \textbf{23.69} & 48.49 & 14.40 & \textbf{17.30}\\ 
 ARIMA/DHR-ARIMA & KmeansPlus.Number & \textbf{13.44} & 12.67 & 11.14 & 34.70 & 23.71 & 48.66 & 14.42 & 17.31\\ 
 & Kmeans.Seed & 13.47 & 12.69 & 11.21 & 34.68 & 23.72 & 48.57 & 14.40 & 17.33\\ 
 & KmeansPlus.Seed & 13.49 & 12.70 & 11.22 & 34.62 & 23.85 & 48.49 & 14.37 & 17.34\\ 
 
 \hline 
\end{tabular}
}
\caption{Mean sMAPE Results of Forecast Combinations Across All Experimental Datasets}
\label{tab:sMAPE_combined_local_global}
\end{center}
\end{table*}

\subsection{Performance of Forecast Combinations with Global and Local Models}

We compare the performance of 6 RNN-based models: RNN Baseline and 5 feature and distance-based clustered ensemble models against the forecast combinations that combine the forecasts of RNN-based models and traditional univariate forecasting models. 
Table \ref{tab:sMAPE_combined_local_global} shows the performance of the forecast combinations compared with the RNN-based models on mean sMAPE.
The performance of all RNN-based models increases after combining them with traditional univariate forecasting models on the M3, M4, Ausgrid half-hourly, Ausgrid monthly, Tourism and Hospital datasets in terms of mean sMAPE. Furthermore, the ensemble model that combines the predictions of DTW.Number, ETS and ARIMA has provided the second-most accurate forecasts on the M4 monthly dataset based on the mean sMAPE, outperforming FFORMA. For the CIF 2016 dataset, forecast combinations show a mixed performance. 
Forecast combinations are substantially outperformed by the RNN-based models on the Kaggle web traffic dataset due to the poor performance of univariate forecasting models.
Overall, the performance of forecast combinations that use 2 univariate forecasting models are better than the combinations that only use a single univariate forecasting model. 

In summary, the results indicate that the accuracy of globally trained models can be further increased by combining them with locally trained univariate models as they incorporate the strengths of both global and local models while mitigating the weaknesses of each other. These forecast combination models contain heterogeneous base models and hence, they are more capable of addressing the heterogeneity in the time series.

\subsection{Computational Performance}

Our experiments are extensive, which is why we have to execute them on different hardware and thus computational times are in general not comparable. To perform a comparison study of computational performance, we execute a subset of the experiments in a controlled environment, namely an Intel(R) Core(TM) i7 processor (2.6GHz) and 32GB of main memory.

Table \ref{exec_times} shows the computational times of twenty models executed in the controlled environment for the Kaggle web traffic dataset. This dataset contains 997 series where each series contains 550 data points. It is the second largest dataset in terms of total data points. Hence, we consider the Kaggle web traffic dataset as a representative dataset to compare the computational times across models.

For each ensemble and non-ensemble clustering model of each baseline GFM, one model is run and its computational time recorded, as we assume that the time taken to cluster is small compared with the time taken to build the models, so that computational times are similar across the clustering variations of the same baseline GFM. The reported computational times of RNN and FFNN based models include hyperparameter tuning time, training time, and forecast calculation time. 

We can see from the table that RNN and PR based models have the highest and lowest computational times, respectively. The ensemble of specialists has the highest computational time across RNN and PR based models whereas Kmeans.Number has the highest computational time across FFNN based models. Traditional univariate benchmarks, DeepAR, N-BEATS and FFNN based models have similar computational times which are higher than the computational time of PR models and lower than the computational time of RNNs. CatBoost and the PR Baseline have similar computational performance. Though the ensembling procedures approximately double the computational complexity, we note that the choice of base model is more important for the computational complexity, with RNNs being approximately 8 times slower than FFNNs, and FFNNs being about 20 times slower than the linear PR approach. Thus, we conclude that the choice of base model is more important than the choice for ensembling, regarding computation time, and ensembling is feasible for all base models considered.

\begin{table}
\centering
\begin{tabular}{lr}
  \hline
Model & Computational Time (Minutes) \\ 
  \hline
ETS & 7 \\ 
  ARIMA & 16 \\ 
  \hline
  FFORMA & 45 \\ 
  CatBoost & 0.39\\
  DeepAR & 9 \\
  N-BEATS & 21 \\
  \hline
  \hline
  RNN Baseline & 68 \\ 
  RNN Kmeans.OC & 65 \\ 
  \hline
  RNN Kmeans.Number & 109 \\ 
  RNN Ensemble.Specialists & 114 \\ 
  RNN Ensemble.Seed & 89 \\ 
  \hline
  \hline
  FFNN Baseline & 9 \\ 
  FFNN Kmeans.OC & 8 \\ 
   \hline
  FFNN Kmeans.Number & 24 \\ 
  FFNN Ensemble.Specialists & 15 \\ 
  FFNN Ensemble.Seed & 14 \\ 
  \hline
  \hline
  PR Baseline & 0.38 \\ 
  PR Kmeans.OC & 0.08 \\ 
  \hline
  PR Kmeans.Number & 0.45 \\ 
  PR Ensemble.Specialists & 5 \\ 
   \hline
\end{tabular}
\caption{Computational times (in minutes) of twenty models for Kaggle web traffic dataset. The benchmark models, the RNN-based models, the FFNN-based models, and the PR-based models are separated with double lines. One computational time is reported per each ensemble and non-ensemble clustering type of each baseline GFM as the computational times of the clustering variations with a given baseline GFM are similar.} 
\label{exec_times}
\end{table}

\subsection{Overall Performance of Ensemble and Non-Ensemble Models}
\begin{sidewaysfigure}
 \centering
   \includegraphics[width=\textwidth]{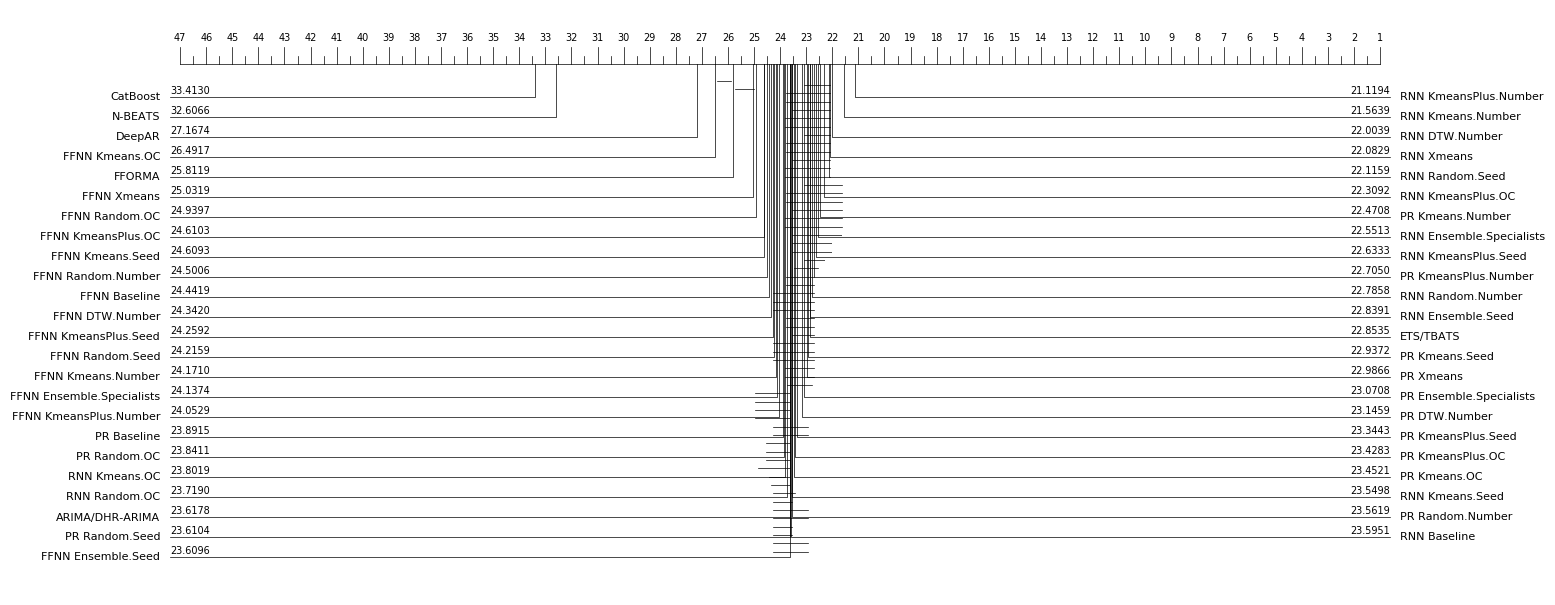}
   \caption{Critical difference diagram showing statistical difference comparison of all considered models. The models are ordered based on their average ranks, which are indicated next to the models, starting from the top right model. The horizontal lines that connect the groups of models indicate that the connected models are not significantly different from each other. Twelve GFMs show a significantly better performance than the traditional univariate benchmarks: ETS, ARIMA, TBATS and DHR-ARIMA.
 }~\label{fig:statistical_test}
 \end{sidewaysfigure}
 
Overall, the proposed ensemble variants outperform the non-ensemble models across all experimental datasets, except on the Kaggle web traffic dataset. 
The performance of the ensemble variants depend on the characteristics of the datasets. The Kaggle web traffic dataset is a homogeneous dataset, where most of the time series share similar characteristics. 
For such datasets, global models that share the same parameters across all series perform best. The clustered ensemble models and the ensemble of specialists perform better on heterogeneous datasets as they group similar series and train different submodels per each subgroup. 
Hence, ensembling is quite helpful in increasing the forecasting accuracy, especially when the time series data are heterogeneous.

Figure \ref{fig:statistical_test} shows a critical difference diagram comparing the models we considered for our experiments. The average ranks are indicated next to the models. The horizontal lines that connect the groups of models indicate that the connected models are not significantly different from each other. Twelve GFMs show a significantly better performance than the traditional univariate benchmarks: ETS, ARIMA, TBATS and DHR-ARIMA. The KmeansPlus.Number cluster variant based on RNNs shows the overall best performance, whereas CatBoost shows the worst performance, out of all considered models. The FFNN-based GFMs overall show a poor performance compared to the RNN and PR-based GFMs. Our ensembled GFMs outperform the benchmarks, based on their average ranks.   

As shown in Figure \ref{fig:statistical_test}, models based on random clustering, namely Random.OC, Random.Number, and Random.Seed, show a better performance compared with the baseline GFMs across PR models. Across RNNs, the Random.Number and Random.Seed models show a better performance whereas across FFNNs, the Random.Seed model shows a better performance compared with the corresponding baseline GFMs. Hence, our results show that increasing the model complexity can improve the forecasting accuracy. When the series are partitioned with more sophisticated clustering techniques to train multiple GFMs, the model complexity gets further increased and the data heterogeneity issues are also properly addressed showing an improved performance over both baseline GFMs and random clustering based GFMs. Ensembling GFMs trained over multiple iterations further increases the model complexity, addresses data, model and parameter uncertainties, and mitigates the information loss that may occur due to the sub-optimal grouping of series. Hence, our clustered ensemble models obtain the top ranks in our evaluation. 

Based on our results, we recommend to use GFM.Cluster.Number approaches over GFM.Cluster.Seed and Ensemble.Specialists approaches to address data heterogeneity (Section \ref{sec:non_clustering_ensembles}). We further recommend to use the distance-based clustering variant, DTW.Number, or the feature-based clustering variant, KmeansPlus.Number, as the base for GFM.Cluster.Number techniques (Section \ref{sec:ensemble_clustering_performance}). As the base GFM, we recommend to use RNNs or PR models over FFNNs (Section \ref{sec:local_global_performance}).

\section{Conclusions and Future Research}
\label{sec:conclusion}

Global forecasting models that are trained across time series have recently become popular in the field of forecasting due to their better performance compared with univariate models that are trained per time series.

In this paper, we have investigated general methodologies to improve the accuracy of GFMs by better localising them, using clustered ensemble models, ensembles of specialists, and ensembles of local and global models, to adequately address data heterogeneity and control model complexity. We have used three base GFMs, namely RNNs, FFNNs, and PR models. 

Across eight publicly available datasets, we have shown that ensembling can improve the forecasting accuracy of GFMs trained over heterogeneous time series compared with the non-ensemble models such as GFM Baselines, clustered non-ensemble models, two deep learning benchmarks: DeepAR and N-BEATS, and four traditional univariate benchmarks: ETS, ARIMA, TBATS, and DHR-ARIMA, .

From our experiments, we conclude that the performance of GFMs can be improved by localising them and by properly addressing data heterogeneity during model training. We further conclude that ensembling is beneficial if a dataset contains diverse time series. Ensembling does not considerably improve the forecasting accuracy for homogeneous datasets, namely the Kaggle web traffic dataset in our experiments, where the GFM Baselines produce better forecasts. The forecast combinations that use both global and local model forecasts have provided the best results over many datasets as they incorporate the strengths of both global and local models while mitigating the weaknesses of each other.

Some areas for further research are as follows. 
In this research, we have only used simple averaging to aggregate the forecasts provided by the localised GFMs. Weighted averaging can be further incorporated to improve the performance of our proposed ensembled localised GFMs. Furthermore, we have only considered point forecasts in a univariate context. Ensembling can be further incorporated to address data uncertainty through probabilistic forecasting. 
Also, many real-world forecasting problems are multivariate, including more than one time-dependent variable and ensembling can be further useful to improve the forecasting accuracy of such forecasting problems.

\section*{Acknowledgements}

This research was supported by the Australian Research Council under grant DE190100045, a Facebook  Statistics  for  Improving  Insights  and  Decisions  research  award,  Monash  University  Graduate Research funding and the MASSIVE High performance computing facility, Australia.

\bibliographystyle{unsrtnat}
\bibliography{references}

\begin{thebibliography}{100}
\providecommand{\natexlab}[1]{#1}
\providecommand{\url}[1]{\texttt{#1}}
\expandafter\ifx\csname urlstyle\endcsname\relax
  \providecommand{\doi}[1]{doi: #1}\else
  \providecommand{\doi}{doi: \begingroup \urlstyle{rm}\Url}\fi

\bibitem[Januschowski et~al.(2020)Januschowski, Gasthaus, Wang, Salinas,
  Flunkert, Bohlke-Schneider, and Callot]{ref_105}
T.~Januschowski, J.~Gasthaus, Y.~Wang, D.~Salinas, V.~Flunkert,
  M.~Bohlke-Schneider, and L.~Callot.
\newblock Criteria for classifying forecasting methods.
\newblock \emph{International Journal of Forecasting}, 36\penalty0
  (1):\penalty0 167--177, 2020.

\bibitem[Makridakis et~al.(2018)Makridakis, Spiliotis, and
  Assimakopoulos]{ref_30}
S.~Makridakis, E.~Spiliotis, and V.~Assimakopoulos.
\newblock The {M4} competition: results, findings, conclusion and way forward.
\newblock \emph{Int. J. Forecast.}, 34\penalty0 (4):\penalty0 802--808, 2018.

\bibitem[Makridakis et~al.(2020)Makridakis, Spiliotis, and
  Assimakopoulos]{makridakis_2020_m5}
S.~Makridakis, E.~Spiliotis, and V.~Assimakopoulos.
\newblock The {M5} accuracy competition: results, findings and conclusions,
  2020.

\bibitem[Hyndman et~al.(2008)Hyndman, Koehler, Ord, and Snyder]{ref_112}
R.~J. Hyndman, A.~B. Koehler, J.~K. Ord, and R.~D. Snyder.
\newblock \emph{Forecasting with Exponential Smoothing: The State Space
  Approach}.
\newblock Springer Science and Business Media, 2008.

\bibitem[Box et~al.(2015)Box, Jenkins, Reinsel, and Ljung]{ref_113}
G.~E.~P. Box, G.~M. Jenkins, G.~C. Reinsel, and G.~M. Ljung.
\newblock \emph{Time Series Analysis: Forecasting and Control}.
\newblock John Wiley and Sons, 2015.

\bibitem[Montero-Manso and Hyndman(2021)]{pablo_2020_principles}
P.~Montero-Manso and R.~J. Hyndman.
\newblock Principles and algorithms for forecasting groups of time series:
  locality and globality.
\newblock \emph{International Journal of Forecasting}, 2021.
\newblock ISSN 0169-2070.

\bibitem[Hewamalage et~al.(2020{\natexlab{a}})Hewamalage, Bergmeir, and
  Bandara]{hewamalage_2020_simulation}
H.~Hewamalage, C.~Bergmeir, and K.~Bandara.
\newblock Global models for time series forecasting: a simulation study.
\newblock \url{https://arxiv.org/abs/2012.12485}, 2020{\natexlab{a}}.

\bibitem[Trapero et~al.(2015)Trapero, Kourentzes, and Fildes]{ref_114}
J.~R. Trapero, N.~Kourentzes, and R.~Fildes.
\newblock On the identification of sales forecasting models in the presence of
  promotions.
\newblock \emph{Journal of the Operational Research Society}, 66\penalty0
  (2):\penalty0 299 -- 307, 2015.

\bibitem[Gelman and Hill(2007)]{gelman2007data}
A.~Gelman and J.~Hill.
\newblock \emph{Data Analysis Using Regression and Multilevel/Hierarchical
  Models}.
\newblock Analytical Methods for Social Research. Cambridge University Press,
  2007.
\newblock ISBN 9780521686891.

\bibitem[Smyl(2020)]{ref_1}
S.~Smyl.
\newblock A hybrid method of exponential smoothing and recurrent neural
  networks for time series forecasting.
\newblock \emph{International Journal of Forecasting}, 36\penalty0
  (1):\penalty0 75--85, 2020.

\bibitem[Sen et~al.(2019)Sen, Yu, and Dhillon]{sen_2019_think}
R.~Sen, H-F Yu, and I.~Dhillon.
\newblock Think globally, act locally: a deep neural network approach to
  high-dimensional time series forecasting.
\newblock In \emph{Advances in Neural Information Processing Systems}, pages
  4837--4846, 2019.

\bibitem[Bandara et~al.(2020{\natexlab{a}})Bandara, Bergmeir, and Smyl]{ref_2}
K.~Bandara, C.~Bergmeir, and S.~Smyl.
\newblock Forecasting across time series databases using recurrent neural
  networks on groups of similar series: a clustering approach.
\newblock \emph{Expert Syst. Appl.}, 140:\penalty0 112896, 2020{\natexlab{a}}.
\newblock ISSN 0957-4174.

\bibitem[Lerch and Baran(2016)]{lerch2016similarity}
S.~Lerch and S.~Baran.
\newblock Similarity‐based semilocal estimation of post‐processing models.
\newblock \emph{Royal Statistical Society}, 66\penalty0 (1):\penalty0 29--51,
  2016.

\bibitem[Godahewa et~al.(2020{\natexlab{a}})Godahewa, Bergmeir, Webb, and
  Montero-Manso]{godahewa_2020_weekly}
R.~Godahewa, C.~Bergmeir, G.~I. Webb, and P.~Montero-Manso.
\newblock A strong baseline for weekly time series forecasting.
\newblock \url{https://arxiv.org/abs/2010.08158}, 2020{\natexlab{a}}.

\bibitem[Schapire(1999)]{schapire_1999_adaboost}
R.~E. Schapire.
\newblock A brief introduction to boosting.
\newblock In \emph{Proceedings of the 16th International Joint Conference on
  Artificial Intelligence - Volume 2}, IJCAI'99, page 1401–1406, San
  Francisco, CA, USA, 1999. Morgan Kaufmann Publishers Inc.

\bibitem[Breiman(2001)]{breiman_2001_random}
L.~Breiman.
\newblock Random forests.
\newblock \emph{Machine Learning}, 45:\penalty0 5--32, 2001.

\bibitem[Bates and Granger(1969)]{granger_1969_combination}
J.~M. Bates and C.~W. Granger.
\newblock The combination of forecasts.
\newblock \emph{Journal of the Operational Research Society}, 20\penalty0
  (4):\penalty0 451––468, 1969.

\bibitem[Timmermann(2006)]{ref_58}
A.~Timmermann.
\newblock Forecast combinations.
\newblock \emph{Handbook of Economic Forecasting}, 1:\penalty0 135–196, 2006.

\bibitem[Wolpert(2002)]{wolpert2002lunch}
D.~H. Wolpert.
\newblock The supervised learning no-free-lunch theorems.
\newblock In R.~Roy, M.~Koppen, S.~Ovaska, T.~Furuhashi, and Hoffmann F.,
  editors, \emph{Soft Computing and Industry}, London, 2002. Springer.

\bibitem[Brown et~al.(2005{\natexlab{a}})Brown, Wyatt, and
  Tino]{brown2005managing}
G.~Brown, J.~L. Wyatt, and P.~Tino.
\newblock Managing diversity in regression ensembles.
\newblock \emph{Journal of Machine Learning Research}, page 1621–1650,
  2005{\natexlab{a}}.

\bibitem[Lloyd(1982)]{ref_90}
S.~P. Lloyd.
\newblock Least squares quantization in {PCM}.
\newblock \emph{Information Theory, IEEE Transactions}, 28\penalty0
  (2):\penalty0 129--137, 1982.

\bibitem[Arthur and Vassilvitskii(2007)]{ref_91}
D.~Arthur and S.~Vassilvitskii.
\newblock K-means++: the advantages of careful seeding.
\newblock In \emph{Proc. of the Annu. ACM-SIAM Symp. on Discrete Algorithms},
  volume~8, pages 1027--1035, 01 2007.

\bibitem[Jin and Han(2010)]{ref_95}
X.~Jin and J.~Han.
\newblock \emph{Encyclopedia of Machine Learning}.
\newblock Springer US, Boston, MA, 2010.
\newblock ISBN 978-0-387-30164-8.

\bibitem[Duncan et~al.(2001)Duncan, Gorr, and
  Szczypula]{duncan_2001_forecasting}
G.~T. Duncan, W.~L. Gorr, and J.~Szczypula.
\newblock \emph{Forecasting Analogous Time Series}, volume~30.
\newblock Springer, Boston, MA, 2001.

\bibitem[Smyl and Kuber(2016)]{ref_23}
S.~Smyl and K.~Kuber.
\newblock Data preprocessing and augmentation for multiple short time series
  forecasting with recurrent neural networks.
\newblock In \emph{36th International Symposium on Forecasting}, 2016.

\bibitem[Salinas et~al.(2020)Salinas, Flunkert, Gasthaus, and
  Januschowski]{ref_99}
D.~Salinas, V.~Flunkert, J.~Gasthaus, and T.~Januschowski.
\newblock {DeepAR}: probabilistic forecasting with autoregressive recurrent
  networks.
\newblock \emph{Int. J. Forecast.}, 36\penalty0 (3):\penalty0 1181--1191, July
  2020.

\bibitem[Oreshkin et~al.(2019)Oreshkin, Carpov, Chapados, and
  Bengio]{oreshkin_2019_nbeats}
B.~N. Oreshkin, D.~Carpov, N.~Chapados, and Y.~Bengio.
\newblock {N-BEATS}: neural basis expansion analysis for interpretable time
  series forecasting.
\newblock \url{https://arxiv.org/abs/1905.10437}, 2019.

\bibitem[Hewamalage et~al.(2020{\natexlab{b}})Hewamalage, Bergmeir, and
  Bandara]{ref_6}
H.~Hewamalage, C.~Bergmeir, and K.~Bandara.
\newblock Recurrent neural networks for time series forecasting: current status
  and future directions.
\newblock \emph{International Journal of Forecasting}, 2020{\natexlab{b}}.
\newblock ISSN 0169-2070.

\bibitem[Bojer and Meldgaard(2020)]{BOJER2020}
C.~S. Bojer and J.~P. Meldgaard.
\newblock Kaggle forecasting competitions: an overlooked learning opportunity.
\newblock \emph{International Journal of Forecasting}, 2020.
\newblock ISSN 0169-2070.

\bibitem[Kent and Hayward(2007)]{kent_2007_limitations}
D.~M. Kent and R.~A. Hayward.
\newblock {Limitations of applying summary results of clinical trials to
  individual patients - the need for risk stratification}.
\newblock \emph{{JAMA}}, 298\penalty0 (10):\penalty0 1209--1212, 09 2007.

\bibitem[Ester et~al.(1996)Ester, Kriegel, Sander, and Xu]{Ester1996ADA}
M.~Ester, H.~Kriegel, J.~Sander, and X.~Xu.
\newblock A density-based algorithm for discovering clusters in large spatial
  databases with noise.
\newblock In \emph{Second International Conference on Knowledge Discovery and
  Data Mining}, page 226–231, 1996.

\bibitem[Kaufman and Rousseeuw(1990{\natexlab{a}})]{kaufman_1990_pam}
L.~Kaufman and P.~J. Rousseeuw.
\newblock \emph{Partitioning Around Medoids (Program PAM)}, pages 68--125.
\newblock John Wiley and Sons, Inc., 1990{\natexlab{a}}.
\newblock ISBN 9780470316801.

\bibitem[Wallace and Dowe(1994)]{wallace1994intrinsic}
C.~S. Wallace and D.~L. Dowe.
\newblock Intrinsic classification by {MML}-the snob program.
\newblock In \emph{Proceedings of the 7th Australian Joint Conference on
  Artificial Intelligence}, volume~37, page~44, 1994.

\bibitem[Bandara et~al.(2019{\natexlab{a}})Bandara, Shi, Bergmeir, Hewamalage,
  Tran, and Seaman]{ref_13}
K.~Bandara, P.~Shi, C.~Bergmeir, H.~Hewamalage, Q.~Tran, and B.~Seaman.
\newblock Sales demand forecast in e-commerce using a long short-term memory
  neural network methodology.
\newblock In \emph{26th International Conference on Neural Information
  Processing}, pages 462--474, 2019{\natexlab{a}}.

\bibitem[Bandara et~al.(2020{\natexlab{b}})Bandara, Bergmeir, Campbell, Scott,
  and Lubman]{Bandara2020-en}
K.~Bandara, C.~Bergmeir, S.~Campbell, D.~Scott, and D.~Lubman.
\newblock Towards accurate predictions and causal 'what-if' analyses for
  planning and policy-making: a case study in emergency medical services
  demand.
\newblock In \emph{International Joint Conference on Neural Networks},
  2020{\natexlab{b}}.

\bibitem[Bandara et~al.(2021)Bandara, Hewamalage, Liu, Kang, and
  Bergmeir]{BANDARA2021108148}
K.~Bandara, H.~Hewamalage, Y.-H. Liu, Y.~Kang, and C.~Bergmeir.
\newblock Improving the accuracy of global forecasting models using time series
  data augmentation.
\newblock \emph{Pattern Recognition}, page 108148, 2021.
\newblock ISSN 0031-3203.

\bibitem[Jose and Winkler(2008)]{ref_59}
V.~R.~R. Jose and R.~L. Winkler.
\newblock Simple robust averages of forecasts: some empirical results.
\newblock \emph{International Journal of Forecasting}, 24\penalty0
  (1):\penalty0 163–169, 2008.

\bibitem[Sanchez(2008)]{ref_62}
I.~Sanchez.
\newblock Adaptive combination of forecasts with application to wind energy.
\newblock \emph{International Journal of Forecasting}, 24\penalty0
  (4):\penalty0 679–693, 2008.

\bibitem[Krogh and Vedelsby(1995)]{krogh_1995_neural}
A.~Krogh and J.~Vedelsby.
\newblock Neural network ensembles, crossvalidation and active learning.
\newblock In \emph{Proceedings of the Advances in Neural Information Processing
  Systems}, pages 231--238, 1995.

\bibitem[Brown et~al.(2005{\natexlab{b}})Brown, Wyatt, Harris, and
  Yao]{brown_2005_diversity}
G.~Brown, J.~L. Wyatt, R.~Harris, and X.~Yao.
\newblock Diversity creation methods: a survey and categorisation.
\newblock \emph{Journal of Information Fusion}, 6\penalty0 (1):\penalty0 5--20,
  2005{\natexlab{b}}.

\bibitem[Cerqueira et~al.(2017)Cerqueira, Torgo, Oliveira, and
  Pfahringer]{ref_51}
V.~Cerqueira, L.~Torgo, M.~Oliveira, and B.~Pfahringer.
\newblock Dynamic and heterogeneous ensembles for time series forecasting.
\newblock In \emph{IEEE International Conference on Data Science and Advanced
  Analytics}, pages 242--251, 2017.

\bibitem[Wolpert(1992)]{ref_52}
D.~H. Wolpert.
\newblock Stacked generalization.
\newblock \emph{Neural Networks}, 5\penalty0 (2):\penalty0 241–259, 1992.

\bibitem[Torgo and Oliveira(2014)]{ref_53}
L.~Torgo and M.~Oliveira.
\newblock Ensembles for time series forecasting.
\newblock In \emph{Asian Conference on Machine Learning}, pages 360--370, 01
  2014.

\bibitem[Heinermann and Kramer(2016)]{ref_55}
J.~Heinermann and O.~Kramer.
\newblock Machine learning ensembles for wind power prediction.
\newblock In \emph{Renewable Energy}, volume~89, pages 671--679, 2016.

\bibitem[Grmanova et~al.(2016)Grmanova, Laurinec, Rozinajova, Ezzeddine, Lucka,
  Lacko, VrablecovÃ¡, and Navrat]{ref_56}
G.~Grmanova, P.~Laurinec, V.~Rozinajova, A.~Ezzeddine, M.~Lucka, P.~Lacko,
  P.~VrablecovÃ¡, and P.~Navrat.
\newblock Incremental ensemble learning for electricity load forecasting.
\newblock \emph{Acta Polytechnica Hungarica}, 13:\penalty0 97--117, 2016.

\bibitem[Ribeiro et~al.(2019)Ribeiro, Ribeiro, Reynoso-Meza, and
  d.~S.~Coelho]{ref_57}
M.~H. D.~M. Ribeiro, V.~H.~A. Ribeiro, G.~Reynoso-Meza, and L.~d.~S.~Coelho.
\newblock Multi-objective ensemble model for short-term price forecasting in
  corn price time series.
\newblock In \emph{International Joint Conference on Neural Networks (IJCNN)},
  pages 1--8, July 2019.

\bibitem[Masoudnia and Ebrahimpour(2014)]{masoudnia2014mixture}
S.~Masoudnia and R.~Ebrahimpour.
\newblock Mixture of experts: a literature survey.
\newblock \emph{Artificial Intelligence Review}, 42\penalty0 (2):\penalty0
  275--293, 2014.

\bibitem[Laurinec et~al.(2019)Laurinec, Loderer, Lucka, and
  Rozinajova]{laurinec2019density}
P.~Laurinec, M.~Loderer, M.~Lucka, and V.~Rozinajova.
\newblock Density-based unsupervised ensemble learning methods for time series
  forecasting of aggregated or clustered electricity consumption.
\newblock \emph{Journal of Intelligent Information Systems}, 53\penalty0
  (2):\penalty0 219--239, 2019.

\bibitem[Montero-Manso et~al.(2020)Montero-Manso, Athanasopoulos, Hyndman, and
  Talagala]{ref_3}
P.~Montero-Manso, G.~Athanasopoulos, R.~J. Hyndman, and T.~S. Talagala.
\newblock {FFORMA}: feature-based forecast model averaging.
\newblock \emph{International Journal of Forecasting}, 36\penalty0
  (1):\penalty0 86--92, 2020.

\bibitem[Pawlikowski and Chorowska(2020)]{PAWLIKOWSKI202093}
M.~Pawlikowski and A.~Chorowska.
\newblock Weighted ensemble of statistical models.
\newblock \emph{International Journal of Forecasting}, 36\penalty0
  (1):\penalty0 93 -- 97, 2020.
\newblock ISSN 0169-2070.
\newblock M4 Competition.

\bibitem[Cerqueira et~al.(2019)Cerqueira, Torgo, Pinto, and
  Soares]{cerqueira2019arbitrage}
V.~Cerqueira, L.~Torgo, F.~Pinto, and C.~Soares.
\newblock Arbitrage of forecasting experts.
\newblock \emph{Machine Learning}, 108\penalty0 (6):\penalty0 913--944, 2019.

\bibitem[Kaufman and Rousseeuw(1990{\natexlab{b}})]{ref_45}
L.~Kaufman and P.~J. Rousseeuw.
\newblock \emph{Finding Groups in Data: An Introduction to Cluster Analysis.}
\newblock John Wiley, 1990{\natexlab{b}}.
\newblock ISBN 978-0-47031680-1.

\bibitem[Warren~Liao(2005)]{ref_44}
T.~Warren~Liao.
\newblock Clustering of time series data — a survey.
\newblock \emph{Pattern Recognit.}, 38\penalty0 (11):\penalty0 1857–1874,
  2005.

\bibitem[Wang et~al.(2006)Wang, Smith, and Hyndman]{ref_46}
X.~Wang, K.~Smith, and R.~J. Hyndman.
\newblock Characteristic-based clustering for time series data.
\newblock \emph{Data Min. Knowl. Discov.}, 13\penalty0 (3):\penalty0 335–364,
  2006.

\bibitem[Fraley and Raftery(2002)]{ref_47}
C.~Fraley and A.~E. Raftery.
\newblock Model-based clustering, discriminant analysis, and density
  estimation.
\newblock \emph{Journal of the American Statistical Association}, 97\penalty0
  (458):\penalty0 611–631, 2002.

\bibitem[Genre et~al.(2013)Genre, Kenny, Meyler, and
  Timmermann]{genre_2013_average}
V.~Genre, G.~Kenny, A.~Meyler, and A.~Timmermann.
\newblock Combining expert forecasts: can anything beat the simple average?
\newblock \emph{International Journal of Forecasting}, 29\penalty0
  (1):\penalty0 108–121, 2013.

\bibitem[Hornik(1991)]{ref_71}
K.~Hornik.
\newblock Approximation capabilities of multilayer feedforward networks.
\newblock \emph{Neural Netw.}, 4\penalty0 (2):\penalty0 251–257, 1991.

\bibitem[Elman(1990)]{ref_8}
J.~L. Elman.
\newblock Finding structure in time.
\newblock \emph{Cogn. Sci.}, 14\penalty0 (2):\penalty0 179--211, 1990.

\bibitem[Goodfellow et~al.(2016)Goodfellow, Bengio, and
  Courville]{Goodfellow-et-al-2016}
I.~Goodfellow, Y.~Bengio, and A.~Courville.
\newblock \emph{Deep Learning}.
\newblock MIT Press, 2016.

\bibitem[Venables and Ripley(2002)]{ref_110}
W.~N. Venables and B.~D. Ripley.
\newblock \emph{Modern Applied Statistics with S}.
\newblock Springer, New York, fourth edition, 2002.

\bibitem[Taieb et~al.(2012)Taieb, Bontempi, Atiya, and
  Sorjamaa]{taieb2012review}
S.~Ben Taieb, G.~Bontempi, A.~F. Atiya, and A.~Sorjamaa.
\newblock A review and comparison of strategies for multi-step ahead time
  series forecasting based on the {NN5} forecasting competition.
\newblock \emph{Expert Systems with Applications}, 39\penalty0 (8):\penalty0
  7067--7083, 2012.

\bibitem[Schafer and Zimmermann(2006)]{ref_7}
A.~M. Schafer and H.~G. Zimmermann.
\newblock Recurrent neural networks are universal approximators.
\newblock In \emph{Artificial Neural Networks – ICANN}, pages 632--640.
  Springer Berlin Heidelberg, 2006.

\bibitem[Bandara et~al.(2019{\natexlab{b}})Bandara, Bergmeir, and
  Hewamalage]{ref_24}
K.~Bandara, C.~Bergmeir, and H.~Hewamalage.
\newblock {LSTM-MSNet}: leveraging forecasts on sets of related time series
  with multiple seasonal patterns.
\newblock \emph{IEEE Transactions on Neural Networks and Learning Systems},
  2019{\natexlab{b}}.

\bibitem[Godahewa et~al.(2020{\natexlab{b}})Godahewa, Deng, Bergmeir, and
  Prouzeau]{ref_12}
R.~Godahewa, C.~Deng, C.~Bergmeir, and A.~Prouzeau.
\newblock Simulation and optimisation of air conditioning systems using machine
  learning.
\newblock \url{https://arxiv.org/abs/2006.15296}, 2020{\natexlab{b}}.

\bibitem[Cho et~al.(2014)Cho, van Merrienboer, Gulcehre, Bahdanau, Bougares,
  Schwenk, and Bengio]{ref_10}
K.~Cho, B.~van Merrienboer, C.~Gulcehre, D.~Bahdanau, F.~Bougares, H.~Schwenk,
  and Y.~Bengio.
\newblock Learning phrase representations using {RNN} encoder–decoder for
  statistical machine translation.
\newblock In \emph{Proceedings of the Conference on Empirical Methods in
  Natural Language Processing (EMNLP)}, pages 1724--1734, 2014.

\bibitem[Abadi et~al.(2015)Abadi, Agarwal, Barham, Brevdo, Chen, Citro,
  Corrado, Davis, Dean, Devin, Ghemawat, Goodfellow, Harp, Irving, Isard, Jia,
  Jozefowicz, Kaiser, Kudlur, Levenberg, Man\'{e}, Monga, Moore, Murray, Olah,
  Schuster, Shlens, Steiner, Sutskever, Talwar, Tucker, Vanhoucke, Vasudevan,
  Vi\'{e}gas, Vinyals, Warden, Wattenberg, Wicke, Yu, and Zheng]{ref_26}
M.~Abadi, A.~Agarwal, P.~Barham, E.~Brevdo, Z.~Chen, C.~Citro, G.~S. Corrado,
  A.~Davis, J.~Dean, M.~Devin, S.~Ghemawat, I.~Goodfellow, A.~Harp, G.~Irving,
  M.~Isard, Y.~Jia, R.~Jozefowicz, L.~Kaiser, M.~Kudlur, J.~Levenberg,
  D.~Man\'{e}, R.~Monga, S.~Moore, D.~Murray, C.~Olah, M.~Schuster, J.~Shlens,
  B.~Steiner, I.~Sutskever, K.~Talwar, P.~Tucker, V.~Vanhoucke, V.~Vasudevan,
  F.~Vi\'{e}gas, O.~Vinyals, P.~Warden, M.~Wattenberg, M.~Wicke, Y.~Yu, and
  X.~Zheng.
\newblock {TensorFlow}: large-scale machine learning on heterogeneous systems,
  2015.
\newblock URL \url{https://www.tensorflow.org/}.
\newblock Software available from tensorflow.org.

\bibitem[Friedman et~al.(2010)Friedman, Hastie, and Tibshirani]{ref_111}
J.~Friedman, T.~Hastie, and R.~Tibshirani.
\newblock Regularization paths for generalized linear models via coordinate
  descent.
\newblock \emph{Journal of Statistical Software}, 33\penalty0 (1):\penalty0
  1--22, 2010.

\bibitem[Cleveland et~al.(1990)Cleveland, Cleveland, McRae, and
  Terpenning]{ref_21}
R.~B. Cleveland, W.~S. Cleveland, J.~E. McRae, and I.~Terpenning.
\newblock {STL}: a seasonal-trend decomposition procedure based on loess.
\newblock \emph{J. Off. Stat.}, 6\penalty0 (1):\penalty0 3--33, 1990.

\bibitem[Hyndman and Athanasopoulos(2018)]{ref_4}
R.~J. Hyndman and G.~Athanasopoulos.
\newblock \emph{{Forecasting: Principles and Practice}}.
\newblock OTexts, 2nd edition, 2018.

\bibitem[Hyndman and Khandakar(2008)]{ref_22}
R.~J. Hyndman and Y.~Khandakar.
\newblock Automatic time series forecasting: the forecast package for {R}.
\newblock \emph{Journal of Statistical Software, Articles}, 27\penalty0
  (3):\penalty0 1--22, 2008.

\bibitem[{R Core Team}(2020)]{stat_2020_pkg}
{R Core Team}.
\newblock \emph{{R}: A Language and Environment for Statistical Computing}.
\newblock R Foundation for Statistical Computing, Vienna, Austria, 2020.

\bibitem[Goerg(2013)]{goerg_2013_licors}
G.~M. Goerg.
\newblock {LICORS}: light cone reconstruction of states - predictive state
  estimation from spatio-temporal data, 2013.
\newblock R package version 0.2.0.

\bibitem[Sardá-Espinosa(2019)]{ref_96}
A.~Sardá-Espinosa.
\newblock Time-series clustering in {R} using the dtwclust package.
\newblock \emph{The R Journal}, 2019.

\bibitem[Hyndman et~al.(2015)Hyndman, Wang, and Laptev]{ref_88}
R.~J. Hyndman, E.~Wang, and N.~Laptev.
\newblock Large-scale unusual time series detection.
\newblock In P.~Cui, J.~Dry, C.~Aggarwal, Z.~Zhou, A.~Tuzhilin, H.~Xiong, and
  X.~Wu, editors, \emph{IEEE International Conference on Data Mining Workshop
  (ICDMW)}, page 1616–1619, United States of America, 2015. IEEE, Institute
  of Electrical and Electronics Engineers.

\bibitem[Hyndman et~al.(2020)Hyndman, Kang, Montero-Manso, Talagala, Wang,
  Yang, and O'Hara-Wild]{ref_89}
R.~J. Hyndman, Y.~Kang, P.~Montero-Manso, T.~Talagala, E.~Wang, Y.~Yang, and
  M.~O'Hara-Wild.
\newblock tsfeatures: time series feature extraction, 2020.
\newblock R package version 1.0.2.9000.

\bibitem[Lubba et~al.(2019)Lubba, Sethi, Knaute, Schultz, Fulcher, and
  Jones]{lubba_2019_catch22}
C.~H. Lubba, S.~S. Sethi, P.~Knaute, S.~R. Schultz, B.~D. Fulcher, and N.~S.
  Jones.
\newblock catch22: canonical time-series characteristics.
\newblock In \emph{Data Min Knowl Disc}, page 1821–1852, 2019.

\bibitem[Christ et~al.(2018)Christ, Braun, Neuffer, and
  Kempa-Liehr]{CHRIST201872}
M.~Christ, N.~Braun, J.~Neuffer, and A.~W. Kempa-Liehr.
\newblock Time series feature extraction on basis of scalable hypothesis tests
  (tsfresh – a python package).
\newblock \emph{Neurocomputing}, 307:\penalty0 72--77, 2018.
\newblock ISSN 0925-2312.

\bibitem[Kassambara and Mundt(2019)]{ref_97}
A.~Kassambara and F.~Mundt.
\newblock factoextra: extract and visualize the results of multivariate data
  analyses. {R} package version 1.0.5.
\newblock \url{https://cran.r-project.org/web/packages/factoextra}, 2019.

\bibitem[Livera et~al.(2011)Livera, Hyndman, and Snyder]{ref_93}
A.~M.~De Livera, R.~J. Hyndman, and R.~Snyder.
\newblock Forecasting time series with complex seasonal patterns using
  exponential smoothing.
\newblock \emph{J. Am. Stat. Assoc.}, 106\penalty0 (496):\penalty0 1513–1527,
  2011.

\bibitem[Hyndman et~al.(2021)Hyndman, Athanasopoulos, Bergmeir, Caceres, Chhay,
  O'Hara-Wild, Petropoulos, Razbash, Wang, and Yasmeen]{ref_94}
R.~J. Hyndman, G.~Athanasopoulos, C.~Bergmeir, G.~Caceres, L.~Chhay,
  M.~O'Hara-Wild, F.~Petropoulos, S.~Razbash, E.~Wang, and F.~Yasmeen.
\newblock {forecast}: forecasting functions for time series and linear models,
  2021.
\newblock {R} package version 8.15.

\bibitem[Makridakis and Hibon(2000)]{ref_5}
S.~Makridakis and M.~Hibon.
\newblock The {M3}-competition: results, conclusions and implications.
\newblock \emph{International Journal of Forecasting}, 16\penalty0
  (4):\penalty0 451–476, 2000.

\bibitem[{\v S}t{\v e}pni{\v c}ka and Burda(2017)]{ref_31}
M.~{\v S}t{\v e}pni{\v c}ka and M.~Burda.
\newblock On the results and observations of the time series forecasting
  competition {CIF} 2016.
\newblock In \emph{2017 {IEEE} International Conference on Fuzzy Systems
  ({FUZZ-IEEE})}, pages 1--6, 2017.

\bibitem[Google(2017)]{ref_32}
Google.
\newblock Web traffic time series forecasting, 2017.
\newblock URL
  \url{https://www.kaggle.com/c/web-traffic-time-series-forecasting}.

\bibitem[AusGrid(2019)]{ref_33}
AusGrid.
\newblock Solar home electricity data, 2019.
\newblock URL
  \url{https://www.ausgrid.com.au/Industry/Our-Research/Data-to-share/Solar-home-electricity-data}.

\bibitem[Athanasopoulos et~al.(2011)Athanasopoulos, Hyndman, Song, and
  Wu]{Athanasopoulos_2011_tourism}
G.~Athanasopoulos, R.~J. Hyndman, H.~Song, and D.~C. Wu.
\newblock The tourism forecasting competition.
\newblock \emph{International Journal of Forecasting}, 27\penalty0
  (3):\penalty0 822--844, 2011.

\bibitem[Hyndman(2015)]{hyndman_2015_expsmooth}
R.~J. Hyndman.
\newblock expsmooth: data sets from forecasting with exponential smoothing.
\newblock \url{https://cran.r-project.org/web/packages/expsmooth}, 2015.

\bibitem[Orabona and Tommasi(2017)]{ref_29}
F.~Orabona and T.~Tommasi.
\newblock Training deep networks without learning rates through coin betting.
\newblock In \emph{Advances in Neural Information Processing Systems},
  volume~30, page 2160–2170, 2017.

\bibitem[Hutter et~al.(2011)Hutter, Hoos, and Leyton-Brown]{ref_34}
F.~Hutter, H.~H. Hoos, and K.~Leyton-Brown.
\newblock Sequential model-based optimization for general algorithm
  configuration.
\newblock In C.~A. Coello, editor, \emph{Learning and Intelligent
  Optimization}, pages 507--523, Berlin, Heidelberg, 2011.

\bibitem[Lindauer et~al.(2017)Lindauer, Eggensperger, Feurer, Falkner,
  Biedenkapp, and Hutter]{ref_35}
M.~Lindauer, K.~Eggensperger, M.~Feurer, S.~Falkner, A.~Biedenkapp, and
  F.~Hutter.
\newblock {SMAC} v3: algorithm configuration in python.
\newblock \url{https://github.com/automl/SMAC3}, 2017.

\bibitem[Hyndman and Koehler(2006)]{ref_36}
R.~J. Hyndman and A.~B. Koehler.
\newblock Another look at measures of forecast accuracy.
\newblock \emph{Int. J. Forecast.}, 22\penalty0 (4):\penalty0 679--688, 2006.

\bibitem[Suilin(2017)]{ref_14}
A.~Suilin.
\newblock kaggle-web-traffic.
\newblock \url{https://github.com/Arturus/kaggle-web-traffic}, 2017.

\bibitem[Alexandrov et~al.(2020)Alexandrov, Benidis, Bohlke-Schneider,
  Flunkert, Gasthaus, Januschowski, Maddix, Rangapuram, Salinas, Schulz,
  Stella, Turkmen, and Wang]{gluonts_2020_alexandrov}
A.~Alexandrov, K.~Benidis, M.~Bohlke-Schneider, V.~Flunkert, J.~Gasthaus,
  T.~Januschowski, D.~C. Maddix, S.~Rangapuram, D.~Salinas, J.~Schulz,
  L.~Stella, A.~C. Turkmen, and Y.~Wang.
\newblock {GluonTS}: probabilistic and neural time series modeling in python.
\newblock \emph{Journal of Machine Learning Research}, 21\penalty0
  (116):\penalty0 1--6, 2020.

\bibitem[Prokhorenkova et~al.(2018)Prokhorenkova, Gusev, Vorobev, Dorogush, and
  Gulin]{NEURIPS2018_14491b75}
L.~Prokhorenkova, G.~Gusev, A.~Vorobev, A.~V. Dorogush, and A.~Gulin.
\newblock Catboost: unbiased boosting with categorical features.
\newblock In S.~Bengio, H.~Wallach, H.~Larochelle, K.~Grauman, N.~Cesa-Bianchi,
  and R.~Garnett, editors, \emph{Advances in Neural Information Processing
  Systems}, volume~31. Curran Associates, Inc., 2018.

\bibitem[Montero-Manso(2020)]{m4metalearning_2020_pablo}
P.~Montero-Manso.
\newblock {M4metalearning}: metalearning tools for time series forecasting,
  2020.
\newblock R package version 0.0.0.9000.

\bibitem[Pelleg and Moore(2000)]{ref_92}
D.~Pelleg and A.~Moore.
\newblock X-means: extending k-means with efficient estimation of the number of
  clusters.
\newblock In \emph{Proceedings of the Seventeenth International Conference on
  Machine Learning}, page 727–734, 2000.

\bibitem[Hornik et~al.(2009)Hornik, Buchta, and Zeileis]{ref_98}
K.~Hornik, C.~Buchta, and A.~Zeileis.
\newblock Open-source machine learning: {R} meets {Weka}.
\newblock \emph{Computational Statistics}, 24\penalty0 (2):\penalty0 225--232,
  2009.

\bibitem[Demsar(2006)]{demsar_2006_statistical}
J.~Demsar.
\newblock Statistical comparisons of classifiers over multiple data sets.
\newblock \emph{Journal of Machine Learning Research}, 7:\penalty0 1--30, 01
  2006.

\bibitem[Friedman(1940)]{friedman_1940_comparison}
M.~Friedman.
\newblock A comparison of alternative tests of significance for the problem of
  m rankings.
\newblock \emph{The Annals of Mathematical Statistics}, 11\penalty0
  (1):\penalty0 86–92, 1940.

\bibitem[Benavoli et~al.(2016)Benavoli, Corani, and
  Mangili]{benavoli_2016_ranks}
A.~Benavoli, G.~Corani, and F.~Mangili.
\newblock Should we really use post-hoc tests based on mean-ranks?
\newblock \emph{Journal of Machine Learning Research}, 17\penalty0
  (1):\penalty0 152–161, 2016.

\bibitem[Fawaz et~al.(2019)Fawaz, Forestier, Weber, Idoumghar, and
  Muller]{IsmailFawaz2018deep}
H.~Fawaz, G.~Forestier, J.~Weber, L.~Idoumghar, and P.~Muller.
\newblock Deep learning for time series classification: a review.
\newblock \emph{Data Mining and Knowledge Discovery}, 33\penalty0 (4):\penalty0
  917--963, 2019.

\end{thebibliography}

\end{document}